\documentclass{article} 
\usepackage{iclr2027_conference,times}

\usepackage{amsmath,amsfonts,bm}

\def\eqref#1{equation~\ref{#1}}

\def\1{\bm{1}}

\DeclareMathAlphabet{\mathsfit}{\encodingdefault}{\sfdefault}{m}{sl}
\SetMathAlphabet{\mathsfit}{bold}{\encodingdefault}{\sfdefault}{bx}{n}

\usepackage{hyperref}
\usepackage{url}

\usepackage{multicol}
\usepackage{tabularx}

\title{\textit{Fantastic Metrics and How to Find Them}: Meta-Evaluating Metrics for Multi-lingual Style-Personalized Generation}

\author{Anubhav Jangra\\ 
Columbia University, USA \\
\texttt{anubhav@cs.columbia.edu}
\And
Adrian de Wynter\\
The University of York, UK \\
\texttt{adrian.dewynter@york.ac.uk}
}

\usepackage{times}
\usepackage{latexsym}
\usepackage[T1]{fontenc}
\usepackage[utf8]{inputenc}
\usepackage{microtype}
\usepackage{inconsolata}
\usepackage{graphicx}

\usepackage{booktabs}      
\usepackage{amsmath}
\usepackage[table]{xcolor} 
\usepackage{subcaption}    
\usepackage{epigraph}
\usepackage{multirow}

\renewcommand{\eqref}[1]{Equation~\ref{eq:#1}}

\DeclareRobustCommand{\DE}[3]{#2}

\usepackage[colorinlistoftodos,prependcaption,textsize=tiny,textwidth=14mm]{todonotes}
\makeatletter
\define@key{todonotes}{AW}[]{%
    \setkeys{todonotes}{author=AW,color=blue!20}}%
\makeatother   
\makeatletter
\define@key{todonotes}{AJ}[]{%
    \setkeys{todonotes}{author=AJ,color=red!20}}%
\makeatother   
\makeatletter
\define@key{todonotes}{MZ}[]{%
    \setkeys{todonotes}{author=MZ,color=green!20}}%
\makeatother

\iclrfinalcopy 
\begin{document}

\maketitle

\begin{abstract}
Meta-evaluation of text generation metrics and LLM-judges is primarily conducted using metric-human correlations. However, prior work has shown that correlation is not an appropriate estimate of a metric's evaluation capability and can be unstable across evaluation settings, such as different datasets or languages. Therefore, we propose a regression-based meta-evaluation of metrics that learns on calibrated metric outputs to predict human ratings. We ground this measurement relative to a trivial constant-mean baseline and human agreements to provide a comparable measure of metric capability across evaluation settings. To test our meta-evaluation framework, we developed a large-scale multi-lingual style-personalized text generation (SPTG) benchmark, collecting multiple annotations across 10 topologically diverse languages and four evaluation dimensions, each with varying degrees of subjectivity. We meta-evaluated 32 metrics from three style evaluation paradigms (string-based, embedding-based, and LLM-judges) and their ensembles over these 40 evaluation settings. Our analysis revealed that even simple string- and embedding-based measurements can match strong LLM-judges when calibrated to human ratings. We also identified characteristics of human annotations that determine metric performance, as well as conditions under which a metric ensemble outperforms its individual constituents. \footnote{Code and data is available at \url{https://github.com/anubhav-jangra/multilingual-sptg-eval}.}

\end{abstract}

\section{Introduction}\label{sec:intro}

Evaluating text generation often requires multiple human judgments for inherently subjective qualities, where there is no single correct answer~\citep{plank2022problem, amidei2019agreement}. As obtaining human annotations is expensive and time-consuming, researchers rely on automatic metrics to validate a system's performance on subjective generation tasks. In standard practice, these automatic metrics are evaluated by comparing their agreement with human ratings, computed as correlation coefficients. This is a widely adopted norm across several tasks including summarization~\citep{fabbri2021summeval, liu2023g}, machine translation~\citep{freitag2021experts}, style transfer~\citep{mir2019evaluating, briakou2021evaluating, ostheimer2024text, mukherjee2025evaluating} and LLM-as-judges~\citep{chiang2023can}. However, correlation gives a poor estimate of a metric's utility for subjective assessments due to three reasons: i) it assumes a consistent relationship between metric and human scores, ii) it is sensitive to ties and label skews~\citep{deutsch2023ties}, and iii) it does not have a reference point for human agreement, making it incomparable across tasks, annotators, and datasets. Prior work has shown that metric correlations can be unstable and misleading~\citep{novikova2017we,mathur2020tangled}. 

As an alternative to correlation-driven meta-evaluation of metrics, we instead propose a regression-based meta-evaluation framework to overcome the above-mentioned limitations. The proposed meta-evaluation approach first calibrates each metric to the human rating scale, and then trains a regressor to predict the human consensus from the metric's output. The quality of the metric is then determined by the trained regressor's performance on held-out data, anchored between a constant predictor and the averaged agreement of human annotators. This anchored quality estimate of the metric is comparable across evaluation settings, and more interpretable than the metric-human correlations. 
We apply this framework to style personalized text generation~\citep{jangra2025evaluating}, a task that requires a model to rewrite a source text in the style of a reference text. We selected this task as it captures several evaluation dimensions with varying subjectivity (factual correctness of source text \textit{vs.} stylistic representation of reference text). As the metric behaviors and annotation properties may vary across languages~\citep{hada2024large, hada2024metal, kocmi2021ship}, we conducted our study in 10 topologically diverse languages. We curated a large-scale multi-lingual dataset for this task, collecting ratings from multiple native speakers across the 10 languages along four evaluation dimensions, namely \textit{factual} accuracy, \textit{style} accuracy, \textit{leakage} and \textit{usability}. We evaluated 32 metrics across a broad spectrum of evaluation paradigms, including string-based metrics, semantic embeddings, style embeddings, and LLM judges. Since the meta-evaluation framework can accept several metrics as inputs, it can evaluate learned combinations of metrics. Our systematic investigation of metric ensembles extends prior work~\citep{jangra2025evaluating} by identifying when and how they improve the evaluation quality over constituent metrics.

Our analysis showed that simple correlation-based assessment hides several indicators of metric quality. Across all four dimensions, the correlation between LLM-judges and human annotations was significantly higher than that of other evaluation measures. However, once calibrated to human ratings, much simpler embeddings and string-based metrics were able to capture style at least as well as the strongest LLM-judges. The meta-evaluation across languages revealed the metric's performance depended on two properties of human annotations: the label skews and annotator agreement. We also showed that metric ensembles improved the most on the style dimension (${\sim}2\mathrm{x}$ improvement over best individual metric) where the constituent metrics were individually weak. Our meta-evaluation revealed no single metric to uniformly perform well across all languages or dimensions. 

Our work makes three key contributions: 1) We proposed a regression-based meta-evaluation framework to evaluate agreement of evaluation metrics on subjective human annotations. 2) We curated a large-scale style-personalized text generation dataset, spanning 10 typologically diverse languages annotated by multiple native speakers across four evaluation dimensions. 3) We conducted an extensive meta-evaluation study over 32 metrics over four evaluation paradigms, including several metric ensembles to examine their performance across different languages and evaluation dimensions. 

\section{Related Work}

\textbf{Meta-evaluation of metrics.} Human evaluations are considered the gold-standard in evaluating natural language generation models~\citep{celikyilmaz2020evaluation}, and are typically validated with their correlation with human ratings~\citep{fabbri2021summeval, mir2019evaluating, chiang2023can, freitag2021experts}. Correlation-based meta-evaluation can be a poor estimate of a metric's utility for subjective human ratings due to their inconsistencies across datasets~\cite{novikova2017we}, sensitivity to outliers~\citep{mathur2020tangled}, high uncertainty~\citep{deutsch2021statistical}, and difficulty processing ties in ranking-based correlations~\citep{deutsch2023ties}. \citet{gehrmann2023repairing} conducted a meta-analysis of 66 language generation papers and found that metrics behave differently across datasets and systems, and that annotator disagreements can reflect genuine task subjectivity rather than noise. They also noted that beyond machine translation, metrics are seldom designed for languages other than English. Our work builds on these limitations in our meta-evaluation, which examines metric performance across four distinct subjective evaluation dimensions across 10 languages, grounding it in annotator agreement and reliability. Regression over human judgments has been previously explored to train new metrics such as \texttt{BLEURT}~\citep{sellam2020bleurt} and \texttt{COMET}~\citep{rei2020comet}, or to combine features to learn an ensemble of metrics~\citep{albrecht2007re, verga2024replacing}. In contrast, we use regression as a means for meta-evaluation, scoring calibrated metrics relative to a mean-label baseline and human agreements.


\textbf{Evaluating style in text generation.} 
We evaluate our work on style-personalized text generation (SPTG), a low-resource-constrained text generation task in which a text generation model rewrites a source text in the style of a reference style grounding~\citep{jangra2025evaluating}. We selected this task because it represents a genuine real-world task in personalized text generation~\citep{salemi2024lamp, kumar2024longlamp} and is inherently subjective. Style transfer has long relied on several measurements to gauge the generation quality, from style strength, content preservation, and fluency~\cite{krishna2020reformulating, protasov2026evaluating, mukherjee2025evaluating, protasov2026evaluating}. As LLMs have become more fluent, we instead focus on four evaluation dimensions - \textit{style accuracy}: does the generated output reflect the reference style, \textit{factual accuracy}: does it adhere to the original text, \textit{leakage}: does it introduce unnecessary content from the style reference, and \textit{usability}: is the generated output a useful rewrite of the original text. These four evaluation dimensions collectively capture a wide spectrum of subjective measures in text generation.

While most prior work in style transfer is limited to English~\citep{jin2022deep}, several works have studied style transfer in multi-lingual settings in specific settings such as formality~\citep{krishna2022few, briakou2021ola} and detoxification~\citep{dementieva2024multiparadetox}. Recent efforts have also developed multi-lingual style-aware embeddings~\citep{qiu2025mstyledistance, kim2025leveraging}, extending their English-only predecessors~\citep{wegmann2022same, patel2024styledistance}. Prior work includes multi-lingual meta-evaluation benchmarks for LLM judges, such as MM-Eval~\citep{son2024mm} and the Cross-Lingual Auto Evaluation suite~\citep{doddapaneni2025cross}. These benchmarks among other prior work~\citep{hada2024large, hada2024metal} showed that the agreement of LLM-judges with human ratings generally drop in low-resource and non-Latin-script languages. However, these benchmarks usually target general response quality or preference, which is usually under-defined in text generation~\citep{gehrmann2023repairing}. These works also limit the analysis to just evaluating LLM-judges, often using correlations or pairwise accuracy. To bridge this gap in prior work, we curated a multi-lingual SPTG benchmark to meta-evaluate metrics on the same task across 10 languages. We also evaluate three other metric paradigms in addition to LLM-judges, showing that, once calibrated, even simpler evaluation metrics can match LLM-judgments on state-of-the-art reasoning models.

\section{Problem Definition}

Let an origin $o$ be a source content document and a reference $r$ be a set of exemplars characterizing a target style. Then a SPTG generator is a model $f:(o,r)\mapsto g$ that produces a candidate $g$ re-expressing the content of $o$ in the style of $r$. For each language $L\in\mathcal{L}$ we are given an annotated corpus $\mathcal{C}_L \;=\; \bigl\{\, (x_{L,i},\, \mathbf{y}_{L,i}) \,\bigr\}_{i=1}^{N_L}$.  Each instance pairs a reference and an origin with the outputs of two candidate generators $f_a,f_b\in\mathcal{F}$,  $x_{L,i} \;=\; \bigl(\, r_{L,i},\; o_{L,i},\; g^{a}_{L,i}, g^{b}_{L,i})$. For each candidate $g\in\{g^{a}_{L,i},g^{b}_{L,i}\}$, multiple humans judge the quality along a set of ordinal evaluation dimensions $e\in\mathcal{E}$, obtaining the ordinal labels $y_{L,i}^{e,j}(g_{L,i}) \in \{0,1, ..., E_e\}$ from annotators $j\in\{1,2,\cdots K\}$. An automatic evaluation metric is a function $m:\;(r,o,g)\;\mapsto\;\mathbb{R}, m\in\mathcal{M},$ that returns a scalar quality estimate for a generation $g$ given its reference and origin. The goal is to find a meta-evaluation functional $\Psi$ that scores how well a metric captures a human evaluation dimension and is comparable across different metrics and languages. Ideally, this meta-evaluation measure must capture how well a metric \textit{m} reproduces the human assessment of a generator $f$'s outputs on dimension $e$ for a language $L$ ($\mathbf{\Psi}^{(e)}_{L}\!\left(m \,;\, f\right)$). Additionally, it should help quantify how much of a human construct $m_p$ recovers that $m_q$ misses in language $L$ ($\Psi^{(e)}_{L}\!\left(m_p\right)-\Psi^{(e)}_{L}\!\left(m_q\right)$). 

\section{Multi-lingual SPTG Dataset Curation} \label{sec:dataset}

The development of our multilingual style-personalized text generation benchmark comprised two stages. First, we translated and corrected the English source texts $(r,o)$ to obtain their corresponding versions in the target languages. Second, these multilingual source texts were used to generate candidate paraphrases, which were subsequently annotated on several evaluation dimensions to construct the final corpus. We describe these two stages in detail in the following subsections.

\subsection{From English corpus to Multi-lingual corpora} \label{subsec:translation}

We started with 1,002 pairs of long-form English text comprising an \textit{origin} text $o$ that is to be re-expressed in the style of a \textit{reference} text $r$. The reference text was obtained from two domains, Reddit \textit{r/ShortStories}\footnote{\url{https://www.kaggle.com/datasets/trevordu/reddit-short-stories}} and Amazon product reviews for Digital Music and Subscription Boxes~\citep{hou2026bridging}, whereas the origin text spanned Reddit \textit{r/ShortStories}, Wikipedia, IMDB, CNN/DailyMail, and Amazon reviews for a diverse content range (refer to Table \ref{tab:english-core}). Using this English corpus as the basis for our multi-lingual SPTG benchmark, we first machine-translated the reference and source texts into the 36 target locales, and then conducted a human-correction study to improve the quality of the translations. Across the 36 locales, native annotators edited the machine translation of most items (typically 80–99\% of the 1,502 items per locale; Table~\ref{tab:translation-stats}). For brevity, provide additional details about the translation study in Appendix~\ref{appx:data-contd}.


\begin{table*}[t]\centering\small\setlength{\tabcolsep}{4pt}
\caption{Dataset and annotation statistics for the curated multi-lingual style personalized text generation benchmark. For each evaluation dimension, we provide the aggregate human label mean (standard deviation), and the ordinal Krippendorff $\alpha$. The arrows indicate the direction for improved generation quality in each dimension. The final row provides an aggregate across all ten languages.}
\label{tab:data-perlang}
\resizebox{0.9\linewidth}{!}{%
\begin{tabular}{lrrrcccccccc}
\toprule
 & \multicolumn{3}{c}{Split} & \multicolumn{2}{c}{Style~$\uparrow$} & \multicolumn{2}{c}{Factual~$\downarrow$} & \multicolumn{2}{c}{Leakage~$\downarrow$} & \multicolumn{2}{c}{Usability~$\downarrow$} \\
\cmidrule(lr){2-4}\cmidrule(lr){5-6}\cmidrule(lr){7-8}\cmidrule(lr){9-10}\cmidrule(lr){11-12}
Language & Train & Dev & Test & mean\,(sd) & $\alpha$ & mean\,(sd) & $\alpha$ & mean\,(sd) & $\alpha$ & mean\,(sd) & $\alpha$ \\
\midrule
Bengali (bn) & 801 & 101 & 100 & 1.24 (1.25) & 0.95 & 0.06 (0.36) & 0.78 & 0.45 (0.83) & 0.98 & 2.02 (1.33) & 0.82 \\
English (en) & 798 & 101 & 103 & 1.52 (0.82) & 0.78 & 0.23 (0.71) & 0.98 & 0.19 (0.53) & 0.99 & 2.75 (0.99) & 0.84 \\
Filipino (fil) & 802 & 101 & 99 & 1.16 (1.37) & 0.58 & 1.51 (1.88) & 0.85 & 0.50 (0.86) & 0.80 & 2.81 (1.42) & 0.58 \\
French (fr) & 802 & 100 & 100 & 1.17 (1.10) & 0.95 & 0.16 (0.59) & 0.68 & 0.54 (0.88) & 0.89 & 2.81 (1.15) & 0.87 \\
Frisian (fy) & 814 & 102 & 102 & 2.28 (1.35) & 0.63 & 0.40 (1.13) & 0.68 & 0.38 (0.78) & 0.91 & 2.32 (1.31) & 0.71 \\
Greek (el) & 802 & 100 & 100 & 1.03 (1.09) & 0.63 & 1.09 (1.73) & 0.74 & 0.55 (0.87) & 0.59 & 2.77 (1.10) & 0.73 \\
Japanese (ja) & 803 & 100 & 101 & 1.30 (1.65) & 0.66 & 0.66 (1.42) & 0.69 & 0.35 (0.73) & 0.78 & 3.31 (1.33) & 0.39 \\
Korean (jo) & 802 & 101 & 99 & 1.48 (1.35) & 0.71 & 0.07 (0.42) & 0.65 & 0.16 (0.54) & 0.62 & 2.34 (1.47) & 0.79 \\
Slovenian (sl) & 801 & 100 & 101 & 1.73 (1.55) & 0.80 & 0.85 (1.63) & 0.97 & 0.42 (0.81) & 0.98 & 2.16 (1.61) & 0.86 \\
Swahili (sw) & 801 & 100 & 101 & 2.42 (1.22) & 0.90 & 0.75 (1.42) & 0.91 & 0.34 (0.72) & 0.92 & 1.45 (1.45) & 0.94 \\
\midrule
Overall & 8,026 & 1,006 & 1,006 & 1.53 (1.37) & 0.76 & 0.58 (1.33) & 0.79 & 0.39 (0.78) & 0.85 & 2.47 (1.42) & 0.75 \\
\bottomrule
\end{tabular}%
}
\end{table*}

\subsection{Style Personalized Text for Individual Language corpus} \label{subsec:style-annotation}

Given the localized reference (style exemplars; $r$) and origin (content; $o$) for each instance, we generate two candidate paraphrases that re-express the origin content in the grounding style, each accompanied by a free-text rationale. The generation models ($f$) are drawn from a nine different models to avoid model-specific biases from any single system \footnote{The candidate generation models included \texttt{dev-gpt-4o-11-20}, \texttt{dev-gpt-45-preview}, \texttt{gpt-41-longco-2025-04-14}, \texttt{gpt-5-mini}, \texttt{gpt-51-2025-11-13}, \texttt{gpt-oss-120b}, \texttt{anthropic-claude-opus-4-1}, \texttt{Phi-4} and \texttt{Qwen-2.5-VL-7B}. The researchers inspected the quality of generations from different models to select only the models that generated coherent outputs for each language, and for some instances, the same model was used to generate both paraphrases as well.}. Obtaining two valid generations per instance enabled us to conduct our SPTG annotation study. For each instance in a language corpus, each candidate paraphrase was independently rated by three human annotators along four ordinal dimensions: \textit{stylistic accuracy} ($0$--$4$; higher $=$ better stylistic alignment with reference), \textit{factual accuracy} ($0$--$4$; higher $=$ more hallucination), \textit{leakage} ($0$--$2$; higher $=$ more information leaked from the reference), and \textit{usability} ($0$--$4$; higher $=$ less usable). Factual accuracy captures content hallucination not present in the generation input $(o,r)$, whereas leakage captures the undesirable content introduced in the generation from the style reference $r$. Besides the paraphrase-level annotations, each rater additionally annotated their pairwise preference between the two outputs (\textit{paraphrase-a}, \textit{paraphrase-b}, \textit{neither} or \textit{both}). Since the annotations for this second phase required much more effort than the translation correction study, we limited the annotations to 10 languages. We leave further investigation into the other languages to future work.

We provide the annotation statistics for all languages in Table \ref{tab:data-perlang}, including the distribution of human labels for the ordinal dimensions as well as the inter-annotator agreements. We observe a clear distinction in the annotations based on the subjectivity of the evaluation dimension. The aggregate values of objective dimensions concentrate on the positive end, with factual ratings ranging from $0.58\pm1.33$ and leakage at $0.39\pm0.78$. On the contrary, the subjective dimension annotations are much more spread out on the ordinal scale, with style scores at $1.53\pm1.37$ and usability at $2.47\pm1.42$. Inter-annotator reliability (evaluated as the mean ordinal Krippendorff's $\alpha$ over languages) is substantial across dimensions and also follows the expected objectivity gradient, with average IAA for leakage (0.85) and factual (0.79) higher than for style (0.76) and usability (0.75). The annotator agreement confirms that agreement on the subjective dimensions is genuine but not uniform either across languages or across specific instances. Our final corpus comprises 10,038 instances, spanning 10 languages, each with two candidate paraphrases, three-annotator ratings on four dimensions, and one side-by-side pairwise preference between the two paraphrases.

\section{Proposed Methodology}

We first describe the statistical diagnostics we conduct to assess the evaluation metrics' capability relative to human performance, following typical evaluation strategies in prior work (Section~\ref{subsec:diagnostics}). Then, in Section~\ref{subsec:meta-metric}, we describe our principled regression-based meta-metric, which provides a normalized score against a trivial baseline and human-annotation performance. 



\subsection{Statistical Diagnostics} \label{subsec:diagnostics}

\textbf{Pointwise alignment.} To identify how good evaluation metrics are aligned with the human judgments; specifically their ordinal pointwise judgments for each of the generated paraphrase, we use Spearman's $\rho$ to measure the metric-human alignment. For an assessment over an evaluation dimension $e$ for language $L$, we aggregate the human labels as the mean of the three annotator ratings, $\bar y^{(e)}_L(g)$, for a candidate paraphrase $g$, to obtain the target rankings for the generations. This correlation is equivalent to a standardized ordinary least squares (OLS) fit over ranked values.


\textbf{Pairwise alignment.} While pointwise evaluation measures how well tuned the metrics are against the human judgments over an evaluation dimension for a language; the pairwise alignment evaluates how good the metrics are at being able to distinguish between two texts within the same context. To evaluate pairwise alignment of a metric with the human annotators, we used the difference in average ordinal ratings of the two paraphrases as the target reference\footnote{Since our pairwise annotations were limited to a single higher-order preference of generation, we chose the difference in ordinal ratings as the pairwise ground truth.}; and used Cohen's $\kappa$ to compute the alignment. Since we computed this alignment on only the decisive instances\footnote{For pairwise instance selection, an annotation pair was considered decisive if the pointwise difference in the aggregate human labels differed by at least 2 points on that dimension, and at least 2 of the 3 annotators individually rank the pair in the same direction as the aggregate labels.}, dropping the instances yielded a smaller subset compared to pointwise evaluation across the four dimensions: style (527), factual (251), leakage (269), and usability (558).


\textbf{Relative agreement across evaluation metrics.} This experiment investigates how much of a human construct does a metric $m_p$ recover, that another metric $m_q$ misses on an evaluation dimension $e$ for language $L$. We fit both metrics jointly against the ranked human labels, and measure the incremental variance $m_p$ explains beyond $m_q$, \textit{i.e.,} the gain in $R^2$ from adding $m_p$ to a model that already contains $m_q$: 
\begin{equation}\label{eq:r-squared}
  \Delta R^2_{p\mid q} \;=\; R^2(m_p, m_q) - R^2(m_q)
\end{equation}

Since ordinary least squares regression finds the best possible linear combination of the metrics, $R^2(m_p, m_q)$ is the highest agreement with human labels that the two metrics can reach together~\citep{cohen2013applied}. Therefore, $\Delta R^2_{p\mid q}$ is the extra agreement that only $m_p$ provides (always $\geq$ 0). It is the squared semi-partial correlation that characterizes the association between $\bar{y}$ and the part of $m_p$ that is uncorrelated with $m_q$. Symmetrically, $\Delta R^2_{q\mid p}$ is $m_q$'s unique contribution, and the remainder $R^2(m_p)+R^2(m_q)-R^2(m_p,m_q)$ is the part shared by both~\citep{seibold1979commonality}. We limit this investigation to pointwise evaluation, for which the metrics assign a score to each item independently, obtaining $m_p$ and $m_q$ on the same per-item units as $\bar{y}$.



\subsection{Meta-metric: Predicting human labels from metric outputs} \label{subsec:meta-metric}

To estimate the capabilities of different evaluation metrics across languages and evaluation dimensions, we propose a principled meta-evaluation of metrics by fitting a regression model to human labels derived from metric predictions. Formally, for a specific language $L$ and evaluation dimension $e$, for each instance, we have two paraphrased generations ${g^a_{L,i}, g^b_{L,i}}$ annotated by three annotators. To train our meta-evaluation regressor, we aggregated the annotations using a reliability weighted strategy to obtain the regression target labels: $\bar{y}^e_{L,i}(g_{L,i}^{a}) = \sum_{j=1}^3w^e_j \text{ . } y^{e,j}_{L,i}(g_{L,i}^a)$, where $w^{e}_{L,j}\ge 0$ reflects annotator $j$'s agreement with the others, and the weights sum to one. Individual annotator reliability was computed using Spearman's correlation with a leave-one-out strategy (Figure~\ref{fig:annotator-reliability}).

Since metrics produce scores on different scales (embeddings between $[-1,1]$, LLM-judges in $\{0,1,\dots E_e\}$), we first map each score onto the human label scale using a monotone calibration, preserving the rankings within the candidates\footnote{LLM-judges in this study are prompted to follow the ordinal scales that were used to collect the human annotation, and therefore are left unchanged in this processing step.}. The meta-metric of any arbitrary metric $m$ is then a regressor trained to predict $\bar{y}$ from $m$'s calibrated output. This arbitrary metric can be a single metric, a specific prompt used for an LLM, or an ensemble of metrics. We evaluated 16 regression models and selected the best-performing regression approach, extremely randomized trees~\citep{geurts2006extremely}, across all the 4$\times$10 dimension-language combinations (see Appendix~\ref{appx:regressor} for further details).

We evaluated each meta-metric with the same 5-fold cross-validation, repeated five times. The folds were split by instance, preventing data contamination from the two generations corresponding to the same instance. The calibration and regression fitting were done only on the training folds, using both candidates per instance. We measured the mean absolute error (MAE) of the held-out predictions. The MAE is not comparable across evaluation settings because it depends on the scale and label skew. Therefore, we normalized the MAE scores on a trivial metric baseline that always predicts the mean label in the training fold ($\mathrm{MAE}_0$), and the oracle error of each annotator in predicting the weighted consensus of the other two, averaged over the three annotators ($\mathrm{MAE}_H$). The absolute metric quality for language $L$ on an evaluation dimension $e$ is thus the fraction of this gap metric $m$ closes: 

\begin{equation}\label{eq:metametric}
\mathbf{\Psi}^{(e)}_{L}(m\,)=
\frac{\mathrm{MAE}_0-\mathrm{MAE}_m}{\mathrm{MAE}_0-\mathrm{MAE}_H},
\end{equation}

A metric quality of 0 indicates that the metric does not add beyond the trivial chance baseline, and 1 means that it agrees with the consensus as well as a single annotator. Note that values above 1 are possible when one annotator is noisier than the consensus. 
\section{Experimental Settings}

\subsection{Off-the-shelf Evaluation Metrics}

We meta-evaluate metrics across four paradigms, ranging from simple string-based overlap to model-based judges. The reference-based metrics (first three paradigms) were computed against each signal's corresponding reference: the style reference $r$ for stylistic accuracy and leakage, and the content source $o$ for factual accuracy. LLM-based judges, on the other hand, were provided with both $r$ and $o$, along with the SPTG output $g$, for all four evaluation dimensions. We also evaluated an ensemble of metrics to assess compositional improvements across diverse metric paradigms (denoted as $m_X$).

\textbf{String-based.} 
These metrics comprise surface-level comparisons, including normalized \texttt{Levenshtein} edit distance~\citep{lcvenshtcin1966binary}, character $3$-gram cosine similarity, verbatim overlap, length ratio, punctuation-rate difference, and token-entropy difference. These metrics are inexpensive, language-agnostic, and cover all ten languages.

\textbf{Text embeddings (semantic and style).}
To measure semantic information overlap, we use the language-agnostic \texttt{BERT} embedding model (\texttt{LaBSE};~\citealp{feng2022language}) as a semantic multi-lingual sentence embedding model. We also used embedding models specifically trained to encode authoring style rather than content, in which the distance in latent space reflects stylistic rather than semantic differences. For English, we used \texttt{Wegmann}~\citep{wegmann2022same} and \texttt{StyleDistance}~\citep{patel2024styledistance} embeddings. We additionally evaluated French (fr-FR), Japanese (ja-JP), and Korean (ko-KR) on its multi-lingual counterpart, \texttt{mStyleDistance}~\citep{qiu2025mstyledistance}, and on French (fr-FR), Japanese (ja-JP) and Greek (el-GR) using another multilingual style representation model, \texttt{mStyleRepr}\citep{kim2025leveraging}. We used cosine distance to measure the semantic and style distance between the SPTG outputs ($g$) and the reference ($r$ or $o$) in the latent space. 

\textbf{LLM-as-a-judge.}
We evaluated a wide spread of models, spanning several proprietary and open-weight LLMs. The proprietary LLMs include \texttt{GPT-5.5} and \texttt{GPT-5.4-mini}~\citep{}, \texttt{Claude-Opus-4.8}, \texttt{Sonnet-5}, and \texttt{Haiku-4.5}~\citep{}, \texttt{Gemini-3.1-Pro} and \texttt{Flash-Lite}~\citep{team2023gemini}; while the opensource models include \texttt{Aya-8B}~\citep{dang2024aya},  \texttt{Gemma-4} (\texttt{12B}, \texttt{31B}, \texttt{E4B})~\citep{team2026gemma}, \texttt{GPT-OSS-20B}~\citep{agarwal2025gpt}, \texttt{Llama-3.1-8B} and \texttt{Llama-3.2-3B}~\citep{grattafiori2024llama}, \texttt{Mistral-7B} and \texttt{Mistral-Nemo-12B}~\citep{}, \texttt{OLMo-3-7B}~\citep{olmo2025olmo}, \texttt{Phi-4-14B}~\citep{abdin2024phi}, \texttt{Qwen-2.5-7B}~\citep{qwen25}, \texttt{Qwen-3-8B}~\citep{yang2025qwen3}, and \texttt{Qwen-3.5} (\texttt{4B/9B/27B})~\citep{team2026qwen3}. Unlike the other reference-based metrics, the judges score all four dimensions, including usability, for which no meaningful reference distance exists. We provide the prompts for pointwise and pairwise evaluations in Appendix~\ref{appx:prompt}. All the open-source models were run on 40Gb NVIDIA A100 GPUs.



\section{Findings}

\subsection{Statistical Diagnostics of Evaluation Metrics}\label{subsec:results-diagnostics}

\begin{table}[t]\centering\small\setlength{\tabcolsep}{4pt}
\caption{Pairwise and pointwise statistical diagnostics over the four evaluation dimensions and their overall aggregate. The evaluation is performed on the test split of the dataset, averaged over human annotations from 10 languages. The pointwise alignment is computed with Spearman's $\rho$, measured against the mean ordinal human ratings for each paraphrase on the respective evaluation dimension. Whereas the pairwise alignment is measured with Cohen's $\kappa$, against the sign of the difference in ordinal ratings for the two paraphrases, dropping the instances where the difference is zero. This table includes only results for the best-performing metrics in our meta-evaluation (Figure~\ref{fig:result-metaeval}); for pointwise and pairwise results across all metrics, refer to Table~\ref{tab:agg-corr}.}
\label{tab:agg-corr-mini}
\resizebox{\linewidth}{!}{%
\begin{tabular}{lccccc|ccccc}\toprule
& \multicolumn{5}{c}{Pointwise (Spearman $\rho$)} & \multicolumn{5}{c}{Pairwise (Cohen's $\kappa$)} \\
\cmidrule(lr){2-6}\cmidrule(lr){7-11}
Metric ($k$) & Style & Factual & Leakage & Usability & Overall & Style & Factual & Leakage & Usability & Overall \\\midrule
\texttt{Levenshtein} & 0.06 & 0.41 & -0.51 & -- & 0.33 & 0.13 & 0.75 & 0.81 & -- & 0.56 \\
\texttt{Verbatim} & 0.03 & -0.38 & 0.55 & -- & 0.32 & -0.01 & 0.69 & 0.87 & -- & 0.52 \\
\hline
\texttt{LaBSE} & 0.07 & 0.41 & -0.57 & -- & 0.35 & 0.11 & 0.76 & 0.85 & -- & 0.57 \\
\hline
\texttt{mStyleRepr} (3) & 0.01 & \textbf{0.50} & -0.58 & -- & 0.36 & -0.11 & 0.83 & 0.80 & -- & 0.50 \\
\hline
\texttt{Gemma4-12B} & \textbf{0.39} & 0.50 & 0.72 & 0.48 & 0.52 & \underline{0.49} & \textbf{0.91} & \underline{0.96} & \underline{0.60} & \underline{0.74} \\
\texttt{Gemma4-31B} & \underline{0.38} & 0.47 & \underline{0.75} & \underline{0.49} & \textbf{0.52} & \textbf{0.52} & \underline{0.88} & \textbf{0.97} & \textbf{0.64} & \textbf{0.75} \\
\hline
\texttt{ClaudeOpus-4.8} & 0.10 & 0.48 & \textbf{0.76} & 0.49 & 0.46 & 0.13 & 0.84 & 0.96 & 0.59 & 0.63 \\
\texttt{GPT-5.5} & 0.07 & 0.45 & 0.73 & \textbf{0.52} & 0.45 & 0.01 & 0.88 & 0.95 & 0.56 & 0.60 \\
\bottomrule\end{tabular}%
}
\end{table}

\textbf{Pointwise alignment}
Table~\ref{tab:agg-corr-mini} reports the Spearman $\rho$ between each metric and the aggregate human label for selective metrics (refer to Table~\ref{tab:agg-corr} for results on all the metrics). The overall alignment between the metrics and the human annotators remained modest, with the best metric achieving a mean correlation of $0.56$. The alignment varied sharply by dimensions, as \textit{leakage} was the most tractable dimension (highest $\rho$=$0.76$) whereas stylistic accuracy was the hardest dimension for every metric (highest $\rho$=$0.40$).

Among the reference-based metrics, \texttt{LaBSE} was the strongest metric (overall $0.35$), slightly outperforming the simpler metrics (\texttt{Levenshtein}, \texttt{Verbatim}, overall ${\sim}0.33$). On the other hand, the style embeddings performed comparatively well on the \textit{style} dimension but collapsed to near-zero correlations with factual accuracy and leakage. For leakage, the reference-based metrics were computed as the distance between the generation $g$ and the content origin $o$ (except \texttt{Verbatim}, which was similarity-based) and therefore showed negative correlations with human ratings. LLM-judges, on the other hand, significantly outperformed the reference-based metrics, with the strongest overall alignment (${\sim}0.53$--$0.56$) obtained by the open-weight \texttt{Gemma4-31B} ($0.52$) and \texttt{GeminiFlashLite-3.1} ($0.53$) for the proprietary LLMs.

\textbf{Pairwise alignment.}
In contrast, the pairwise agreements were higher than the pointwise alignment across the board. However, the comparison between the two evaluation settings is not fair, as the pointwise evaluation was conducted over the entire test set, whereas the pairwise comparison only spanned the instances with a distinctly superior paraphrase, encompassing $\sim250$-$530$ instances across the dimensions. Nonetheless, the pairwise alignment trends broadly aligned with the pointwise results. The ordering of the evaluation dimensions was preserved from the pointwise evaluation and was even sharper. Leakage and factual accuracy were close to the human ceiling for strong LLM-judges ($\kappa \geq 0.90$), whereas stylistic accuracy remained the hardest dimension for both views ($\kappa \leq 0.52$).


\begin{figure}
    \centering
    \includegraphics[width=0.8\linewidth]{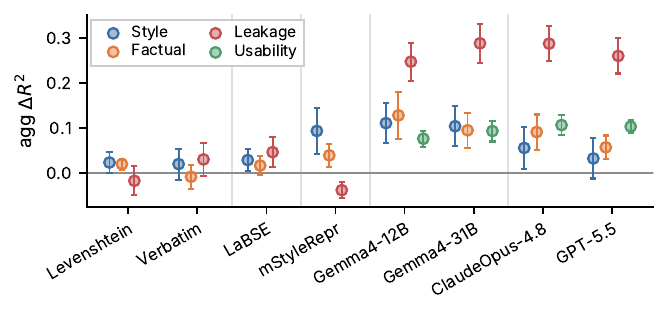}
    \caption{The aggregate $\Delta R^2_{p|.}$ (Eq.~\ref{eq:r-squared}) for selective evaluation metrics over the four evaluation dimensions against the rest of the evaluation metrics (excluding style embeddings due to their limited language). For the results on all the metrics, refer to Figure~\ref{fig:relative-diagnostics} in Appendix~\ref{appx:results}.} \label{fig:relative-diagnostics-mini}
\end{figure}

\textbf{Inter-metric variance decomposition.} 
Beyond absolute alignment against human annotators, we investigated how much unique information each metric contributes relative to other metrics. In Figure~\ref{fig:relative-diagnostics-mini}, we report the average $\Delta R^2$ for the corresponding metrics in Table~\ref{tab:agg-corr-mini} across all dimensions, capturing the variance it adds relative to the pool of all other metrics. Only the LLM-judge metrics had a net positive contribution across all dimensions, while the surface-level string-based metrics and smaller open models were net redundant, providing negligible additional information to the metric pool. Aligning with the pointwise and pairwise alignment trends, the spread is widest for the leakage dimension (up to $+0.35$) and near-collapse on stylistic accuracy across all metrics, indicating that style is a dimension on which metrics differ least in the unique information they carry. Comparing the variance across metric paradigms, we observed that proprietary judges added substantially over every other paradigm ($+0.22$ to $0.37$ in leakage), while the reverse added minimally ($\sim+0.02$). Among the non-LLM metrics, \texttt{LaBSE} stood out as an exception in the leakage dimension, where it added a non-trivial $+0.20$ over the open-weight judges. We provide the pairwise $\Delta R^2_{m_p|m_q}$ and paradigm-level aggregates in Appendix~\ref{appx:results} (Figures~\ref{fig:per-paradigm-relative} and~\ref{fig:full-relative}). 


\subsection{Meta-evaluation of metrics}\label{subsec:results-single-metric}

\begin{figure}
    \centering
    \includegraphics[width=0.9\linewidth]{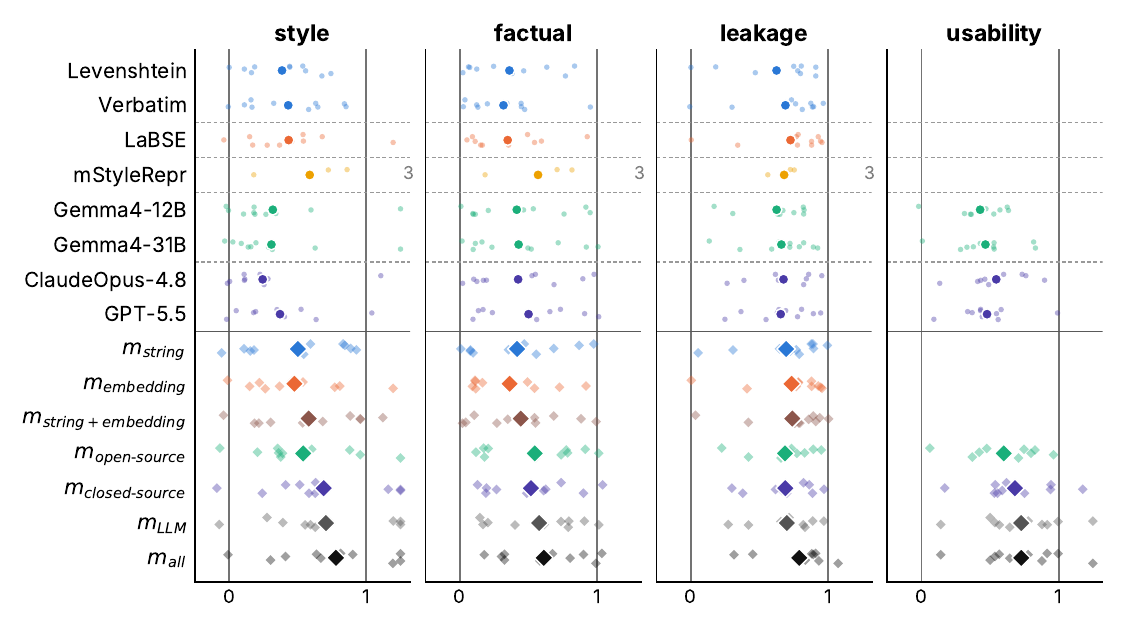}
    \caption{Meta-evaluation of metrics evaluated using the normalized quality score $\Psi$ (Eq.~\ref{eq:metametric}), averaged over 10 languages. $\Psi^e=0$ indicates the chance performance, whereas $\Psi=1$ corresponds to the average human performance. We report the results for the best-performing individual metrics in the corresponding evaluation paradigm as well as several combinations of ensemble metrics ($\rho$). We provide the detailed results for all the metrics in Tables~\ref{tab:q-single} and~\ref{tab:q-perlang} in Appendix~\ref{appx:results}.}
    \label{fig:result-metaeval}
\end{figure}

\textbf{How does calibration affect the performance of individual metrics?} 
Our meta-evaluation revealed that the metric performance varied significantly across evaluation dimensions (Figure~\ref{fig:result-metaeval}). Across all evaluation dimensions, best-performing open-weight models performed equivalently to proprietary models, whereas the small open models scored up to $\Psi=0.15$ on average (\textit{e.g.,} \texttt{Aya-8B} and \texttt{Mistral-7B}). On content-based evaluation dimensions (\textit{factual} and \textit{leakage}), reference-based metrics performed competitively against the best-performing LLM-judges. Whereas learning \textit{style} accuracy from a single LLM-judge output was the hardest, even the strongest LLMs underperformed simpler string-based metrics and \texttt{LaBSE}. This finding contradicts the metric rankings obtained from the aggregate correlation of \texttt{LaBSE} in Table~\ref{tab:agg-corr-mini}. While the per-language correlations partially explain the gap, as \texttt{LaBSE} correlates positively with style in most languages but strongly negatively in Frisian and Japanese, canceling out its signal in the averaged results (Fig.~\ref{fig:perlang-pointwise}, top left); general trends in correlation still indicated it underperformed competitive LLM-judges. As the calibration for each language-dimension pair learns each metric's direction and scale separately, the meta-evaluation recovers the signal to assess the metric's predictive capabilities. While this shows that reference-based metrics can match LLM judgments across the evaluated dimensions, they still require human annotations to calibrate the measurements for general applicability.

\textbf{How do annotations affect meta-evaluation across languages?}
For individual metrics, the performance varied at least as much across languages as across different metrics. Overall, English and French scored the lowest, while Japanese, Greek, and Frisian scored highest (Table~\ref{tab:q-perlang}). These trends could not be justified by a language's resource availability for model training, nor by its scripts. We found two properties of human labels that explained most of this variation. \textbf{Firstly}, the spread of the labels per dimension determined the prediction space. For the factual and leakage dimensions, $\Psi$ was the lowest, with most candidates receiving the same ratings. For instance, the factual and leakage scores for Bengali, Korean, French, and English were lower across all metrics, as their labels were skewed toward the best rating for 84-94\% of candidates. \textbf{Secondly}, because we normalized the MAE loss of the meta-evaluation regressor to the averaged inter-annotator predictability (Section~\ref{subsec:meta-metric}), the label spread determined how strict the human reference was for the normalized scores. On style and usability, $\Psi$ was lower for languages where the annotators agreed more closely. There was a strong negative correlation (Spearman $\rho=-0.71$) between the human agreement (Krippendorff's $\alpha$; Table~\ref{tab:data-perlang}) and the normalized scores. English performance was the worst on all four dimensions because it combined both effects, with the highest annotator agreement and the least label spread. Metrics performed poorly on factual consistency and leakage in Korean, even though the annotators had the lowest agreement and the most skewed labels, which relaxed the gap between the human and trivial baselines for normalization. Conversely, $\Psi_{\text{ja}}^{usability}$ was above one, as one annotator (A2) had high disagreement with the other two (Figure~\ref{fig:annotator-reliability}), implying that the metrics matched the annotator consensus more closely than one of the three annotators and not because the evaluation in the language was easy.

\textbf{When are ensemble metrics most effective?}
Every metric ensemble outperformed its constituents, with the largest gains observed for the weakest evaluation dimension, style accuracy (Figure~\ref{fig:result-metaeval}). Ensemble of all metrics obtained $\Psi^{style}(\rho_{all}) = 0.78$, almost twice as much as the best performing single metric $\Psi^{style}(\texttt{LaBSE} \text{ or } \texttt{Verbatim}) = 0.43$. Usability showed a similar but less pronounced trend. On content-based evaluation dimensions (factual consistency and leakage), where the individual metrics already performed better, ensembling metrics did not yield substantial gains. We identified from our variance decomposition diagnostics (Figure~\ref{fig:relative-diagnostics}) that no individual metric contributes more than ${\sim}0.1$ unique variance on style. However, combining them yielded the largest ensemble gain. This indicates that the style signal is thinly spread across multiple metrics, and can only be recovered by combining them. In contrast, for content-based dimensions, this signal is concentrated in a few metrics; hence, ensembling them does not yield significant gains. Examining the ensemble constituents for $m_{all}$ also aligned with this trend, as a single metric typically contributed more than all other metrics (Figure~\ref{fig:ensemble-contrib-factual-usability}, Appendix~\ref{appx:results}). This is more pronounced in leakage, where the ensemble effectively behaves similarly to a single constituent metric, whereas for the style dimension, the contribution is comparatively more spread out, leading to significant improvements over individual metrics. For all languages, no constituent contributes universally to the ensemble's performance, except for usability, where the ensemble relies on the same judges more often than others. 

\section{Conclusion}

In this work, we proposed a regression-based meta-evaluation to assess metrics for subjective tasks, such as style evaluation. For the meta-evaluation, we curated a large-scale multi-lingual dataset for style-personalized text generation, collecting annotations from multiple native speakers across 10 languages along four evaluation dimensions. We argued against using uncalibrated metric correlations with human ratings and showed that raw correlations masked several indicators of metric quality. We analyzed the meta-metric across languages and found its assessment to depend on two properties of human annotations: their label skews and annotator agreement. Finally, we examined the role constituent metrics play within metric ensembles. Ensembles improved most on style, the dimension where the signal was spread thinly across many metrics. The best-performing metrics varied across languages, with no single metric important for all languages or evaluation dimensions. Together, these findings suggest that metric evaluation for subjective text generation should align annotation properties with metric scores, and that learnt combinations of metrics offer a more faithful measure of metric capability.

\subsection*{AI use statement}

We have used generative AI tools to aid in writing the code to conduct the experiments, to serve as synthetic data generators, and to generate the figures and tables. We have not used generative AI for ideation, literature review, or end-to-end paper or section writing. We have reviewed all AI-assisted work, conducted extensive human evaluation studies to improve the quality of synthetic data, validated the code used for experimentation, cross-checked the illustration artifacts, and ensured that all authors had full agency over the paper-writing process. We take responsibility for the final content of this work, including text, claims, or artifacts produced with the aid of generative AI.

\DeclareRobustCommand{\DE}[3]{#3}
\bibliography{main}
\bibliographystyle{iclr2027_conference}

\DeclareRobustCommand{\DE}[3]{#2} 

\clearpage

\appendix

\section{Selecting the regressor for meta-evaluation}\label{appx:regressor}

Our formulation of the meta-metric (Eq.~\ref{eq:metametric}) is applicable to any regression model, and to select one for our meta-metric evaluation, we compared 16 models across 4 families. All metrics were trained on the same calibrated model outputs and evaluated on the same folds (as described in Section~\ref{subsec:meta-metric}) over the test split of the dataset comprising roughly 100 instances (Table~\ref{tab:data-perlang}). The \textbf{linear} regression models comprised of least squares (\texttt{LR})~\citep{hastie2009elements}, ridge with a fixed or cross-validated penalty (\texttt{Ridge}, \texttt{Ridge-CV})~\citep{hoerl1970ridge, golub1979generalized}, \texttt{Lasso}~\citep{tibshirani1996regression}, \texttt{ElasticNet}~\citep{zou2005regularization}, Huber-loss regression (\texttt{Huber})~\citep{huber1992robust}, and two logit-link models that rescale the label to $(0,1)$ and fit it on the logit scale (\texttt{LogitLink}, \texttt{LogitLink-OLS})~\citep{papke1996econometric}. \textbf{Kernel and instance-based} models included support-vector machines and kernel ridge regression with an RBF kernel (\texttt{SVR}, \texttt{KernelRidge})~\citep{drucker1996support,saunders1998ridge}, and distance-weighted $k$-nearest neighbors (\texttt{kNN})~\citep{Altman01081992}. \textbf{Tree ensembles} models included \texttt{RandomForest}~\citep{breiman2001random}, extremely randomized trees (\texttt{ExtraTrees})~\citep{geurts2006extremely}, and \texttt{GradientBoosting}~\citep{friedman2001greedy}. We also evaluated simple \textbf{neural models} multilayer perceptrons with one or two hidden layers (\texttt{MLP-1}, \texttt{MLP-2})~\citep{rumelhart1986learning}.

\begin{figure}[h]
    \centering
    \includegraphics[width=0.8\linewidth]{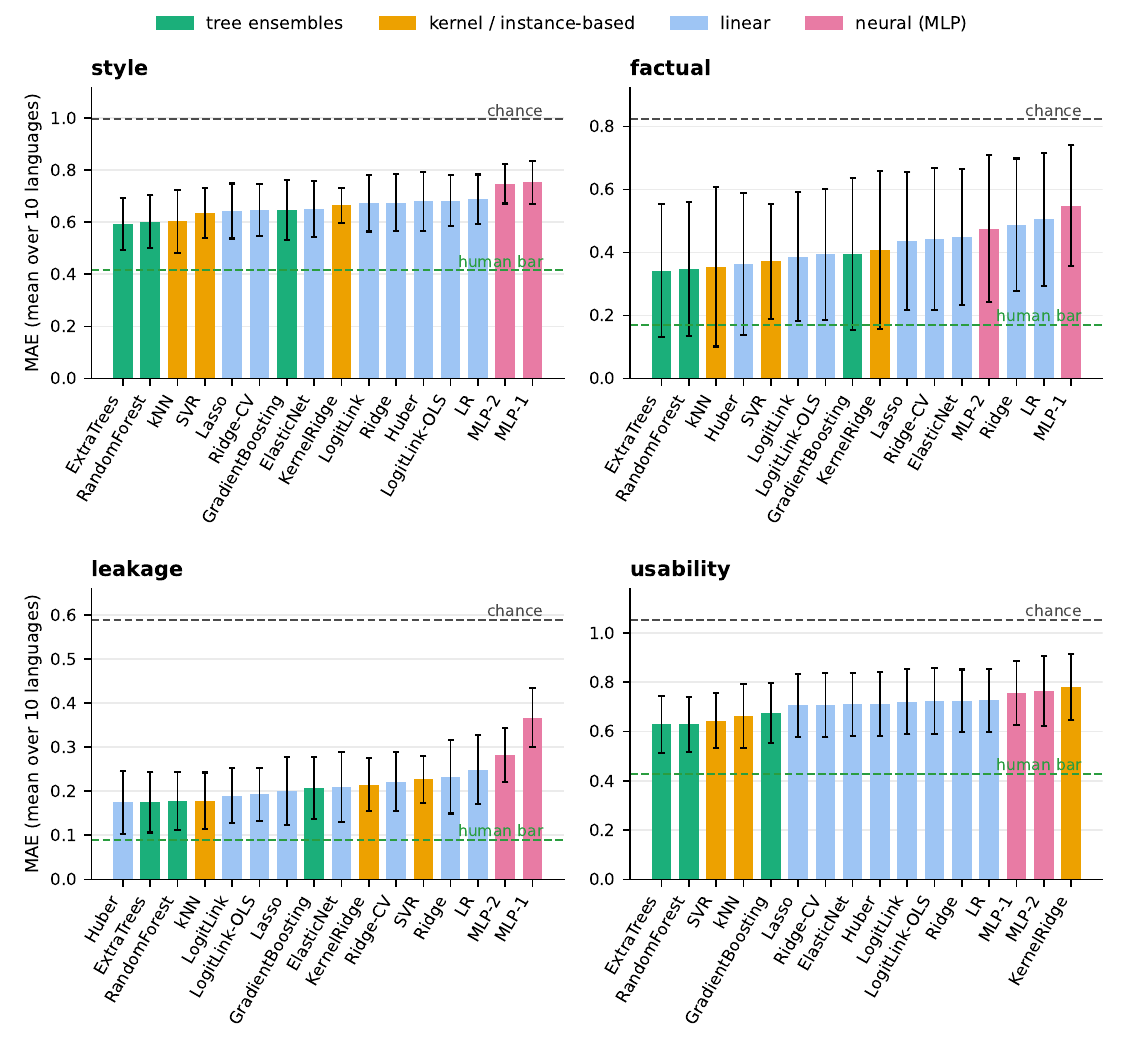}
    \caption{Performance of 16 regression models spanning four regression approaches. The regression models were     trained to fit the metric outputs from all 32 evaluation metrics (\textit{i.e.,} the $\rho_{all}$ ensemble) to predict the reliability-weighted human annotations in a 5-fold cross-validation repeated five times on 10 languages for four evaluation dimensions (\textit{style} accuracy, \textit{factual} accuracy, \textit{leaks} and \textit{usability}).}
    \label{fig:regressor}
\end{figure}

Figure~\ref{fig:regressor} shows the mean absolute error (MAE) of all 16 regression models. Overall, tree ensemble-based approaches outperformed other regressors, with \texttt{ExtraTrees} and \texttt{RandomForest} achieving the lowest mean ranks across languages and evaluation dimensions (2.4 and 2.9). Due to the small size of the training features (32 metrics) and training data (${\sim}$160 generations for 80 instances), the simple least-squares approach overfitted the training data (rank 13.2), with regularized models performing better (rank$_\texttt{Lasso}=$ 7.0). Neural MLP models were among the weakest models (rank$_{\texttt{MLP}_1}=$15.5 and rank$_{\texttt{MLP}_2}=$ 14.5). We selected the best-ranking model, \texttt{ExtraTrees} (300 trees, minimum leaf size 3), as the regression model for the meta-evaluation in this work.
\section{Dataset Curation (Additional Details)} \label{appx:data-contd}



To extend the SPTG corpus to other languages, we sampled 1,002 pairs of long-form English text, comprising an original content text and a reference-style text (Table~\ref{tab:english-core}). The raw translations were then reviewed by a native-speaking annotator, who corrected the accuracy, fluency, and locale-appropriateness of the output, with an optional note explaining their edits. After this step, every instance consequently had three versions of a text $t$ for a language $L$: the standard English source $t_{en}$, the raw machine translation output $t_{L}^{MT}$, and the human-corrected translation $t_{L}^{hc}$. 

The amount of human corrections varied widely across locales (Table~\ref{tab:translation-stats}), from under 1\% of characters edited in five locales to 33.9\% for French, 23.2\% for German, and ${\sim}$22\% for Zulu and Vietnamese, with the large majority of translations altered in most locales (e.g.\ 97-99\% of items edited in French, Vietnamese, and Turkish). Annotators left 5,461 free-text notes documenting their edits, mostly in Maltese, Welsh, Albanian and Irish. Categorizing these notes surfaces six recurring types of corrections: \textit{fluency}, \textit{lexical}, \textit{grammatical}, \textit{spelling}, \textit{mistranslation}, \ textit {dialect-specific}, and others (see Table \ref{tab:correction-types} for examples).

\begin{figure}
    \centering
    \includegraphics[width=0.8\linewidth]{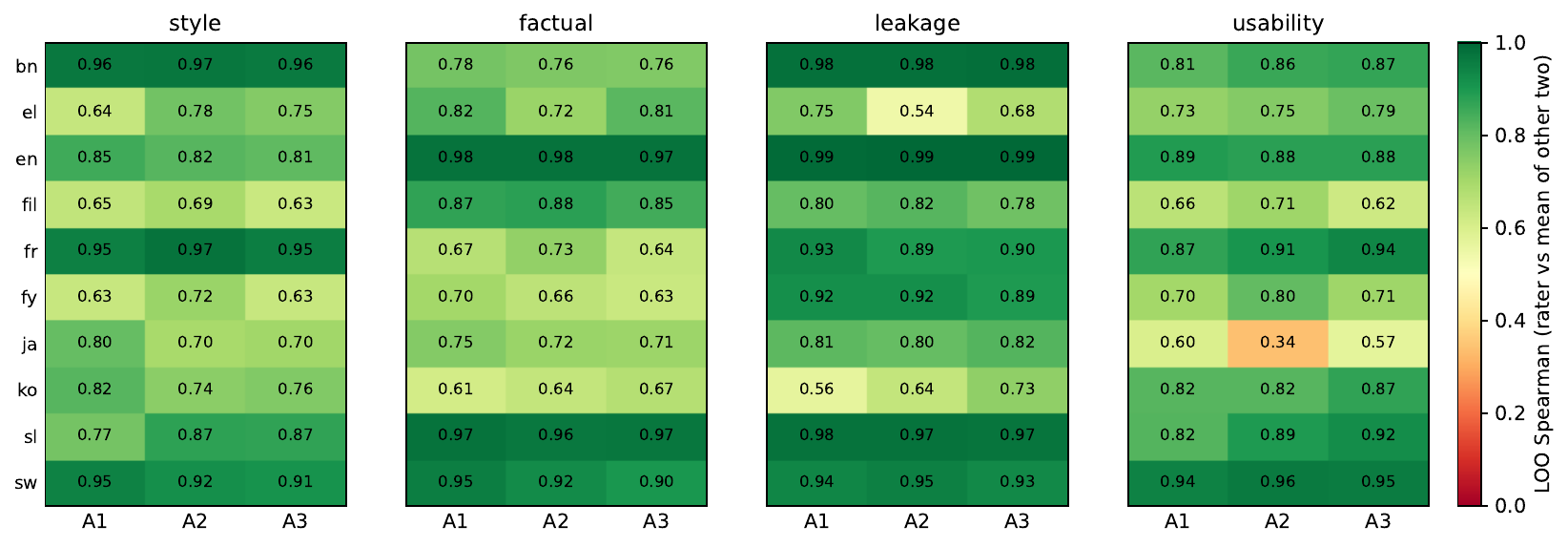}
    \caption{Per-language annotator reliability for all three annotators computed using Spearman's correlation with a leave-one-out strategy, \textit{i.e.,} computing correlation between one annotator and the mean rating from the other two.}
    \label{fig:annotator-reliability}
\end{figure}

We also provide the annotator reliability for each annotator across evaluation dimensions and languages in Figure~\ref{fig:annotator-reliability}. This leave-one-out correlation between one annotator and the mean rating of the other two was used as the reliability weight for each annotator in our meta-metric implementation (Section~\ref{subsec:meta-metric}).


\begin{table}[t]\centering\small
\caption{Composition of the English corpus from which the multilingual SPTG corpus is built ($N{=}1{,}002$ annotated instances, drawn from the $1{,}502$ translated source instances). Each instance pairs a reference (a set of style exemplars) with an independently sampled origin (the content to be re-expressed), and the two are matched at random. References are drawn mostly from creative short-form writing (r/ShortStories; \url{https://www.kaggle.com/datasets/trevordu/reddit-short-stories}), with the remainder from Amazon product reviews~\citep{hou2026bridging}; origins span fiction, encyclopedic text (Wikipedia), movie reviews (IMDB;~\citealp{maas2011learning}), product reviews, and news (CNN/DailyMail;~\citealp{hermann2015teaching}).}
\label{tab:english-core}
\begin{tabular}{lrr}
\toprule
Composition of the English corpus & Count & \% \\
\midrule
\textbf{Reference (style source) domain} & & \\
\quad r/ShortStories & 857 & 85.5 \\
\quad Amazon Reviews (Digital Music) & 121 & 12.1 \\
\quad Amazon Reviews (Subscription Boxes) & 24 & 2.4 \\
\midrule
\textbf{Origin (content source) domain} & & \\
\quad r/ShortStories & 341 & 34.0 \\
\quad Wikipedia & 320 & 31.9 \\
\quad IMDB & 146 & 14.6 \\
\quad Amazon Reviews (Digital Music) & 129 & 12.9 \\
\quad CNN/DailyMail & 34 & 3.4 \\
\quad Amazon Reviews (Subscription Boxes) & 32 & 3.2 \\
\textbf{Total instances} & \textbf{1,002} & \textbf{100.0} \\
\bottomrule
\end{tabular}
\end{table}

\begin{table}[t]\centering\small
\caption{Post-editing statistics for LLM-generated translations. The native speakers were given the original English and model-translated outputs to make corrections. For each locale, we report the percentage of translations that the post-editors changed (\emph{Edited}), the extent of these changes as the character-level edit distance between the machine translation and its corrected version, normalized by length (\emph{Post-edit}; mean $\pm$ s.d.\ in \%), and the total number of free-text comments left by the annotators (\emph{Notes}). Locales are sorted by post-edit distance.}
\label{tab:translation-stats}
\resizebox{0.6\linewidth}{!}{%
\begin{tabular}{lrrcr}
\toprule
Language & Items & Edited (\%) & Post-edit \% ($\mu\pm\sigma$) & Notes \\
\midrule
English (en-US) & 1,002 & -- & -- & -- \\
\midrule
French (fr-FR) & 1,002 & 98 & 36.2 $\pm$ 9.7 & 0 \\
Zulu (zu) & 1,002 & 76 & 32.8 $\pm$ 24.3 & 0 \\
German (de) & 1,002 & 92 & 27.0 $\pm$ 14.9 & 1 \\
Maltese (mt) & 1,002 & 94 & 26.5 $\pm$ 20.5 & 1,805 \\
Danish (da) & 1,002 & 96 & 17.0 $\pm$ 12.5 & 5 \\
Estonian (et) & 1,002 & 94 & 16.2 $\pm$ 23.4 & 7 \\
Norwegian (nb-NO) & 1,002 & 93 & 15.9 $\pm$ 13.5 & 4 \\
Turkish (tr) & 1,002 & 99 & 14.2 $\pm$ 15.7 & 8 \\
Vietnamese (vi) & 1,002 & 98 & 11.0 $\pm$ 19.0 & 0 \\
Malay (ms) & 1,002 & 89 & 6.8 $\pm$ 11.7 & 8 \\
Catalan (ca) & 1,002 & 89 & 5.7 $\pm$ 9.4 & 0 \\
Slovak (sk) & 1,002 & 96 & 5.3 $\pm$ 10.9 & 7 \\
Swahili (sw-SW) & 1,002 & 66 & 4.8 $\pm$ 12.2 & 7 \\
Chinese (Simpl.) (zh-Hans-Cn) & 1,002 & 99 & 4.4 $\pm$ 4.4 & 0 \\
Irish (ga) & 1,002 & 88 & 4.1 $\pm$ 7.8 & 216 \\
Thai (th) & 1,002 & 100 & 3.9 $\pm$ 4.9 & 5 \\
Quechua (qu) & 1,002 & 100 & 3.1 $\pm$ 5.3 & 1 \\
Polish (pl) & 1,002 & 97 & 2.9 $\pm$ 6.0 & 7 \\
Icelandic (is-IS) & 1,002 & 98 & 2.8 $\pm$ 4.6 & 10 \\
Korean (ko-KR) & 1,002 & 79 & 2.6 $\pm$ 5.9 & 6 \\
Greek (el-GR) & 1,002 & 86 & 2.5 $\pm$ 3.3 & 9 \\
Welsh (cy) & 1,002 & 97 & 2.3 $\pm$ 3.7 & 582 \\
Chinese (Trad.) (zh-Hant) & 1,002 & 96 & 1.9 $\pm$ 3.3 & 0 \\
Finnish (fi) & 1,002 & 83 & 1.5 $\pm$ 6.2 & 6 \\
Arabic (ar) & 1,002 & 100 & 1.5 $\pm$ 3.7 & 0 \\
Filipino (fil-PH) & 1,002 & 92 & 1.5 $\pm$ 5.3 & 11 \\
Hebrew (he-IL) & 1,002 & 83 & 1.4 $\pm$ 2.8 & 11 \\
Russian (ru) & 1,002 & 86 & 1.3 $\pm$ 3.3 & 5 \\
Slovenian (sl-SI) & 1,002 & 83 & 1.2 $\pm$ 3.5 & 8 \\
Japanese (ja-JP) & 1,002 & 81 & 1.2 $\pm$ 3.8 & 6 \\
Frisian (fy-NL) & 1,002 & 96 & 1.1 $\pm$ 1.4 & 10 \\
Dutch (nl-NL) & 1,002 & 94 & 1.1 $\pm$ 1.8 & 12 \\
Hindi (hi-IN) & 1,002 & 85 & 1.1 $\pm$ 4.2 & 58 \\
Albanian (sq) & 1,002 & 73 & 1.0 $\pm$ 3.8 & 442 \\
Kazakh (kk) & 1,002 & 24 & 0.5 $\pm$ 1.8 & 2 \\
Spanish (MX) (es-MX) & 1,002 & 61 & 0.5 $\pm$ 1.0 & 6 \\
\bottomrule
\end{tabular}
}
\end{table}

\begin{table*}[t]\centering\small
\caption{Types of corrections noted by post-editors, with examples from different locales. We assigned each free-text note to one type by keyword matching (e.g.\ \emph{typo}$\to$Spelling, \emph{tense}$\to$Grammar, \emph{register}$\to$Dialect). Most notes were too short to classify and fall under \emph{Other}, so the breakdown is only indicative. The number of notes also varies widely by locale (Maltese post-editors wrote most of them), reflecting differences in annotation habits as much as in translation quality.}
\label{tab:correction-types}
\resizebox{0.8\linewidth}{!}{%
\begin{tabularx}{\linewidth}{@{}l r X@{}}
\toprule
Correction type & Notes (\%) & Representative annotator notes \\
\midrule
Fluency / phrasing &
576 \; (11) &
\textbf{[Irish]} rephrased to \emph{i gcuid mhaith}, which ``sounds more natural''; \quad
\textbf{[Maltese]} removed a duplicated \emph{li} in the second sentence; \quad
\textbf{[Hebrew]} first sentence ``feels odd both in English and in Hebrew''. \\

Lexical / word choice &
400 \; (7) &
\textbf{[Irish]} ``NOLA'' expanded to ``New Orleans''; \quad
\textbf{[Welsh]} \emph{(g)wraig} (``wife'') replaced by \emph{menyw} (``woman''). \\

Grammar / morphology &
232 \; (4) &
\textbf{[Welsh]} \emph{agosawu} corrected to \emph{agos} (``close''); \quad
\textbf{[Maltese]} verb aspect \emph{ninhass}$\rightarrow$\emph{qed inxomm}, past tense \emph{insista}$\rightarrow$\emph{insistejt}. \\

Spelling / typo &
218 \; (4) &
\textbf{[Welsh]} \emph{Stifft} corrected to \emph{stiff}; \quad
\textbf{[Welsh]} corrected the spelling of ``Volkswagen''. \\

Mistranslation / wrong sense &
121 \; (2) &
\textbf{[Welsh]} ``crackling'' rendered as \emph{cyfarth} (``barking''), corrected to \emph{cracio}; \quad
\textbf{[Welsh]} \emph{dioddefaint} (``suffering'') used where \emph{poen} (``pain'') was meant; \quad
\textbf{[Hindi]} ``Coordinated Universal Time'' left untranslated. \\

Dialect / register &
109 \; (2) &
\textbf{[Welsh]} \emph{llefrith} (``milk'', North Wales) vs.\ \emph{llaeth} (South Wales); \quad
\textbf{[Maltese]} removed the informal filler \emph{mela KOLLOX}; \quad
\textbf{[Albanian]} passage written in the Geg dialect. \\
\midrule
Other / unspecified &
3{,}805 \; (70) &
Too terse to classify, e.g.\ ``Whirling dervish corrected'', ``Minor word correction''. \\
\midrule
\textbf{Total notes} & \textbf{5{,}461} \; (100) & \\
\bottomrule
\end{tabularx}
}
\end{table*}





\section{Extended Results}\label{appx:results}

\begin{figure}
    \centering
    \includegraphics[width=\linewidth]{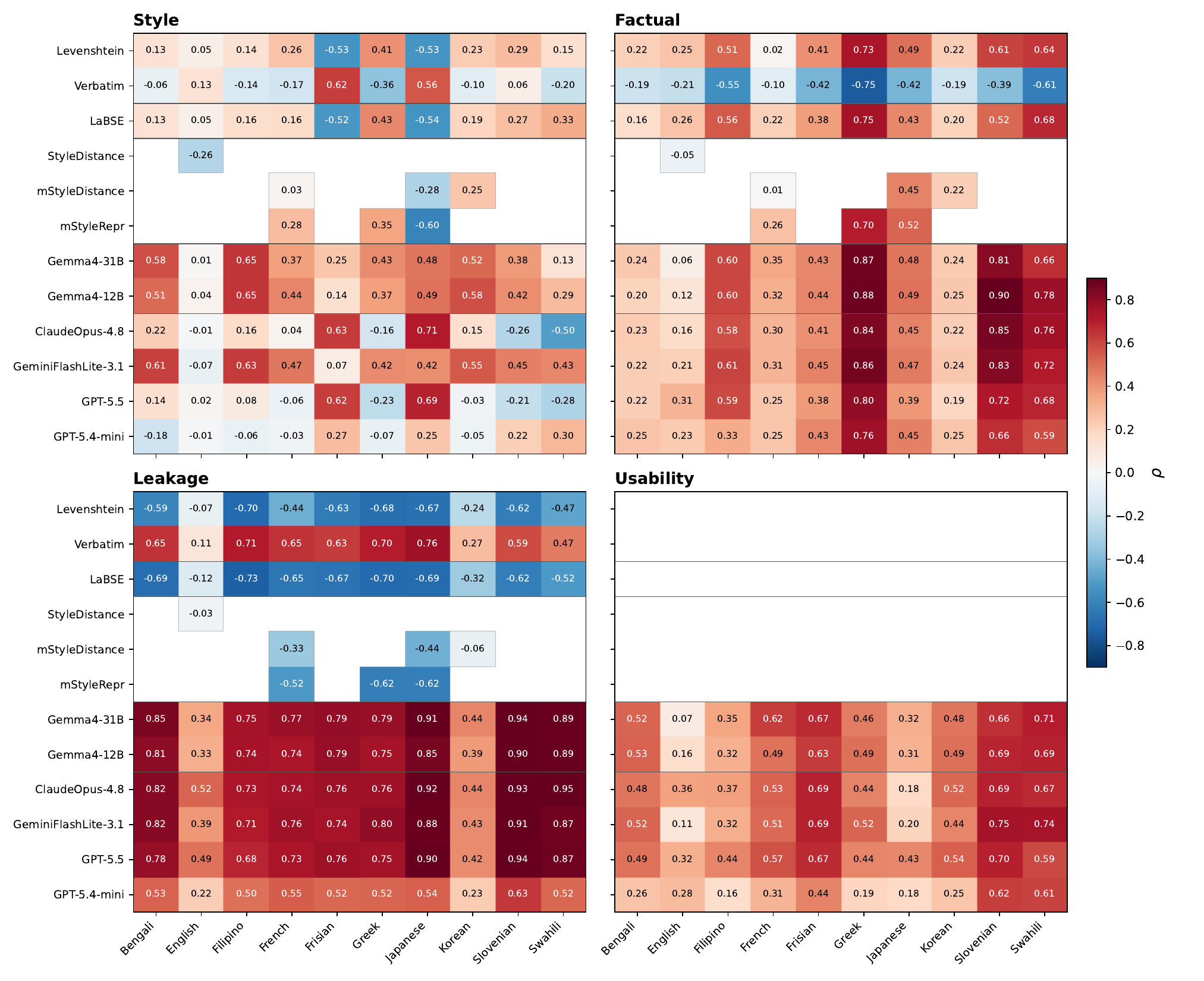}
    \caption{Per-language Spearman's $\rho$ for the pointwise evaluation of the top-2 overall best-performing metrics for each paradigm for each evaluation dimension.}
    \label{fig:perlang-pointwise}
\end{figure}

\begin{figure}
    \centering
    \includegraphics[width=\linewidth]{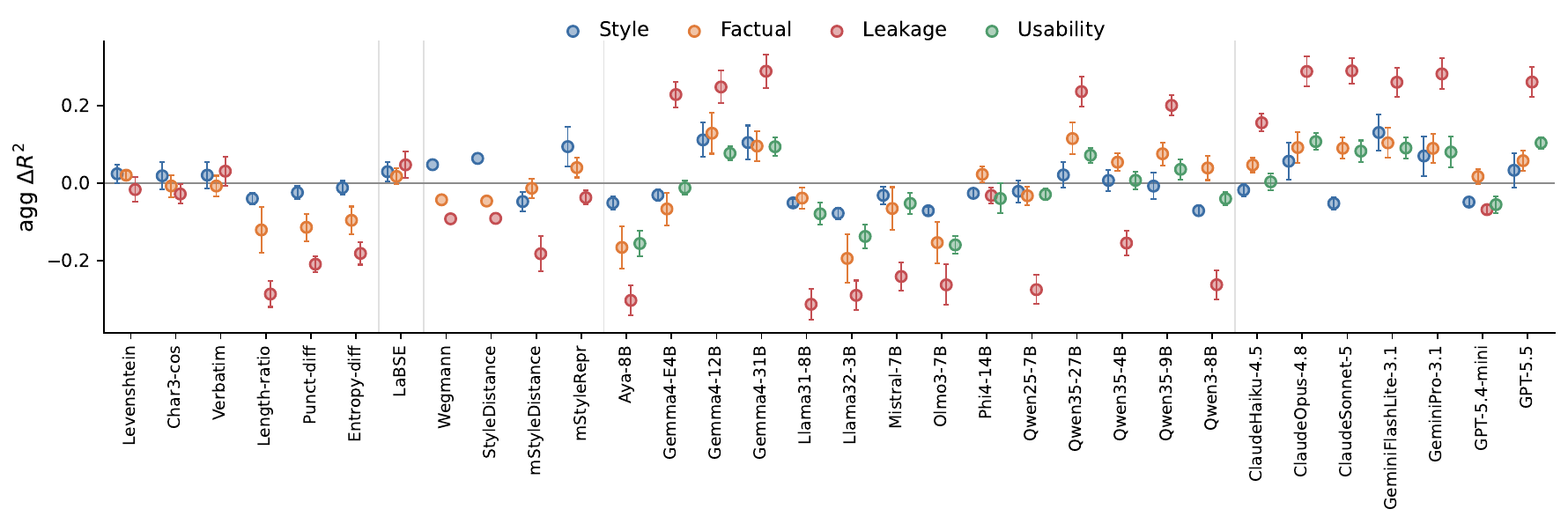}
    \caption{The aggregate $\Delta R^2_{p|.}$ (Eq.~\ref{eq:r-squared}) for all evaluation metrics over the four evaluation dimensions against the rest of the evaluation metrics (excluding style embeddings due to their limited language).}
    \label{fig:relative-diagnostics}
\end{figure}

\begin{figure}
    \centering
    \includegraphics[width=\linewidth]{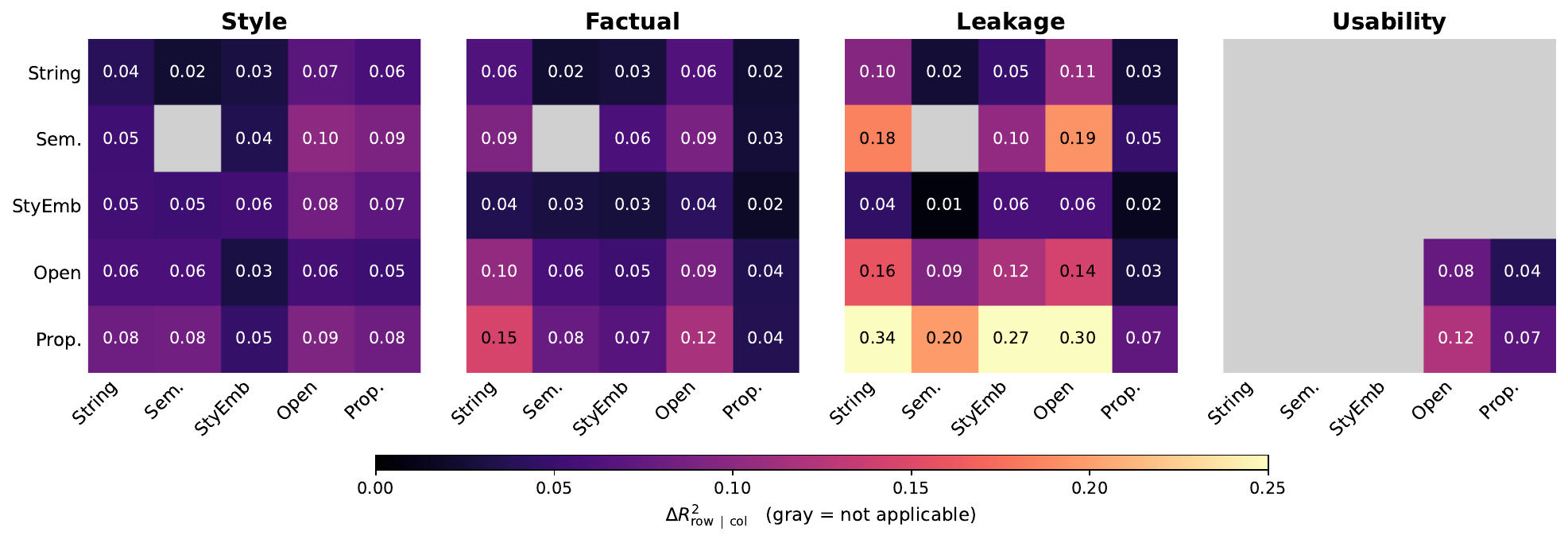}
    \caption{Average $\Delta R^2_{p\mid q}$ (Eq.~\ref{eq:r-squared}) over different evaluation metric paradigms. Each entry $(x,y)$ denotes the average $\frac{1}{|x| + |y|}\sum_{x_i}\sum_{y_j}\Delta R^2_{x_i\mid y_j}$ for evaluation paradigms  $x$ and $y$. The semantic embeddings paradigm has only one metric, so we leave the corresponding cell blank. Similarly, usability is evaluated only with LLM-specific metrics; hence, we leave the other comparisons blank.}
    \label{fig:per-paradigm-relative}
\end{figure}

\begin{figure}
    \centering
    \includegraphics[width=\linewidth]{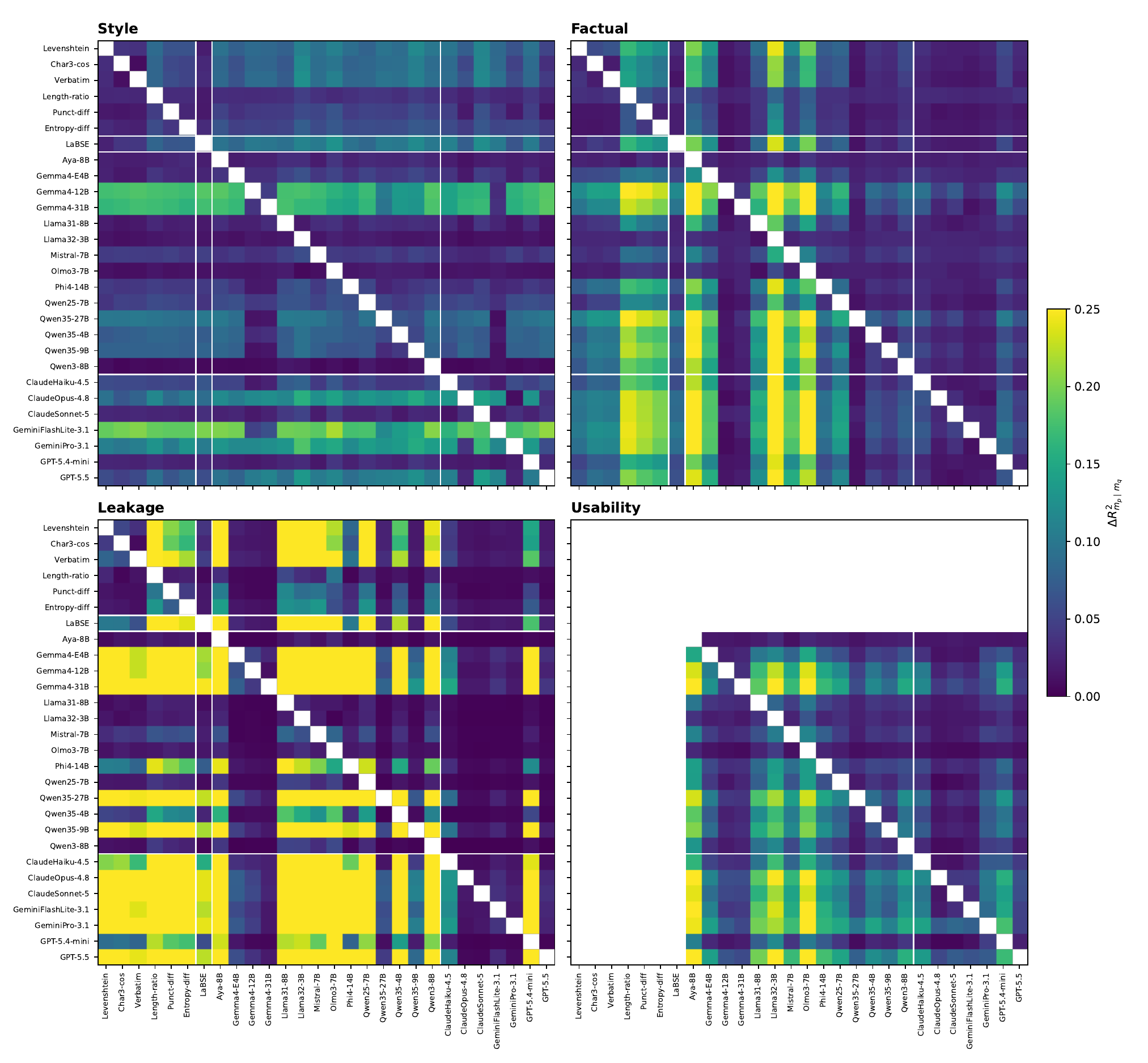}
    \caption{$\Delta R^2_{p\mid q}$ (Eq.~\ref{eq:r-squared}) for a metric pair $m_p, m_q$. The brighter the cell, the more information it provides relative to the other metrics in the corresponding evaluation dimension.}
    \label{fig:full-relative}
\end{figure}

We provide the extended results from our meta-evaluation study in this section. Table~\ref{tab:agg-corr} presents the pointwise and pairwise correlations of all the evaluation metrics. Figure~\ref{fig:perlang-pointwise} shows the per-language Spearman correlation for the pointwise evaluation on the best-performing metrics. Figures~\ref{fig:relative-diagnostics},~\ref{fig:per-paradigm-relative} and~\ref{fig:full-relative} provide the extensive inter-metric variation decomposition using the $\Delta R^2_{p|.}$ (Eq.~\ref{eq:r-squared}) for each metric, across different evaluation paradigms and for all individual metrics respectively.

\begin{figure}
    \centering
    \includegraphics[width=0.95\linewidth]{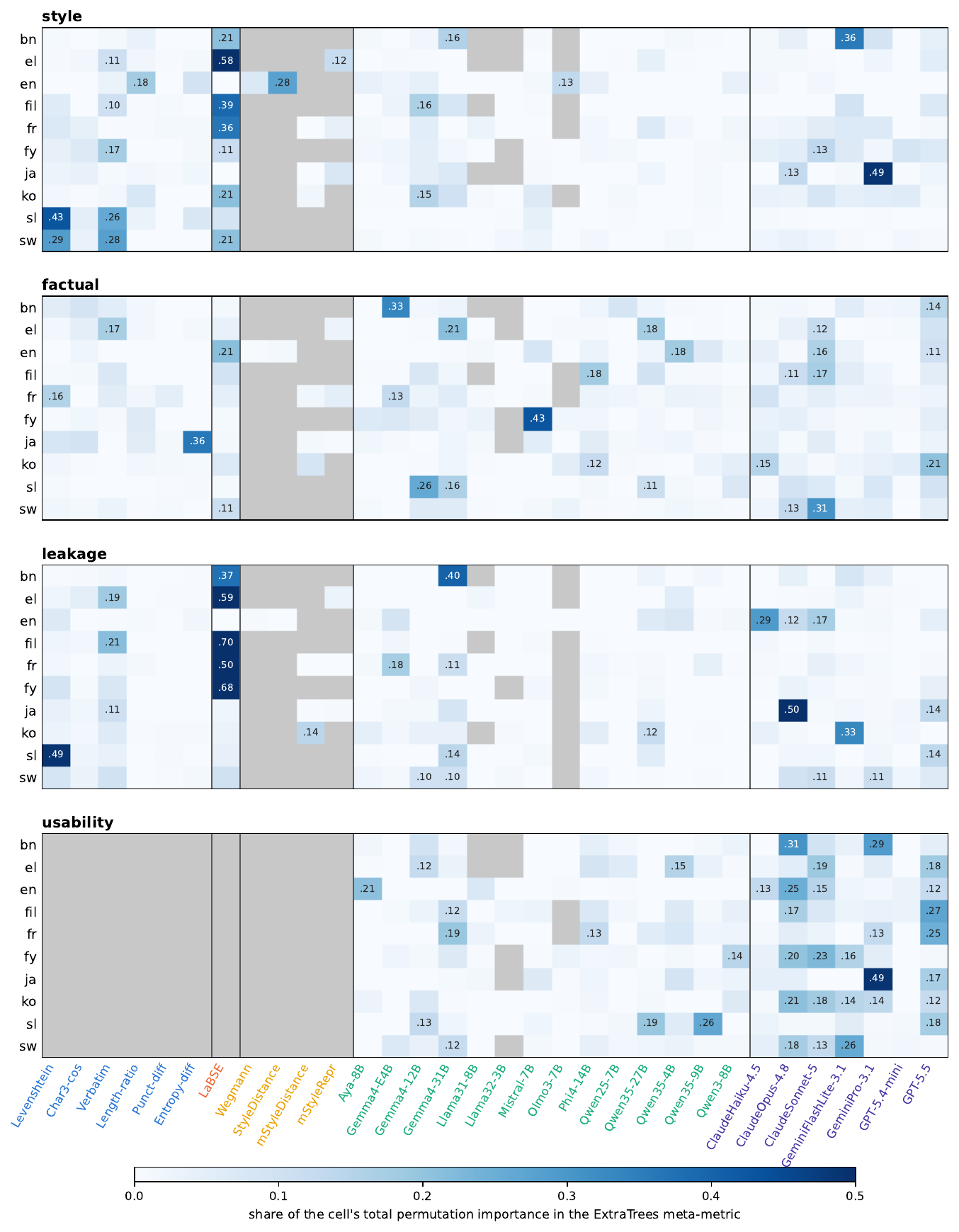}
    \caption{Contribution of each metric to the full ensemble $m_{\text{all}}$ for all four evaluation dimensions on the 10 languages. Each cell is the metric's share of the ensemble's out-of-fold permutation importance, computed as the metric's normalized importance. The importance of a metric is computed as the increase in out-of-fold MAE, computed thrice over all 25 out-folds. Negative importance values were clipped to zero before normalizing to obtain the final contribution of each constituent metric. Gray cells indicate that the metric is unavailable for that language. LLMs with less than half of the valid test set judgments are dropped out (see Table~\ref{tab:parse-errors-perlang}).}
    \label{fig:ensemble-contrib-factual-usability}
\end{figure}

Figure~\ref{fig:ensemble-contrib-factual-usability} illustrates the contribution of all constituent metrics for the $m_{all}$ ensemble metric across the 40 evaluation dimension and language combinations. We provide the aggregate results over all 10 languages from our meta-evaluation analysis (Eq.~\ref{eq:metametric}) in Table~\ref{tab:q-single}, and the extensive per-language results in Table~\ref{tab:q-perlang}. Finally, we provide the parsing error distribution for all LLM-judge outputs for pointwise and pairwise metric generations in Table~\ref{tab:parse-errors-perlang}. 

\begin{table}[t]\centering\small\setlength{\tabcolsep}{4pt}
\caption{Pairwise and pointwise statistical diagnostics of evaluation metrics over the four evaluation dimensions and their overall aggregate (mean $\rho$ for pointwise, and mean $\kappa$ for pairwise). The evaluation is performed on the test split of the dataset, averaged over human annotations from 10 languages. The pointwise alignment is computed with Spearman's $\rho$, measured against the mean ordinal human ratings for each paraphrase on the respective evaluation dimension. Whereas the pairwise alignment is measured with Cohen's $\kappa$, against the sign of the difference in ordinal ratings for the two paraphrases, dropping the instances where the difference is zero.}
\label{tab:agg-corr}
\resizebox{0.85\linewidth}{!}{%
\begin{tabular}{lccccc|ccccc}
\toprule
& \multicolumn{5}{c}{Pointwise (Spearman $\rho$)} & \multicolumn{5}{c}{Pairwise (Cohen's $\kappa$)} \\
\cmidrule(lr){2-6}\cmidrule(lr){7-11}
Metric ($k$) & Style & Factual & Leaks & Usability & Overall & Style & Factual & Leaks & Usability & Overall \\
\midrule
\texttt{Levenshtein} & \textbf{0.06} & \textbf{0.41} & -0.51 & -- & \textbf{0.33} & \textbf{0.13} & \textbf{0.75} & 0.81 & -- & \textbf{0.56} \\
\texttt{Char3-cos} & -0.02 & \underline{0.38} & -0.50 & -- & 0.30 & 0.03 & 0.66 & \underline{0.81} & -- & 0.50 \\
\texttt{Verbatim} & 0.03 & -0.38 & \textbf{0.55} & -- & \underline{0.32} & -0.01 & \underline{0.69} & \textbf{0.87} & -- & \underline{0.52} \\
\texttt{Length-ratio} & \underline{0.05} & 0.17 & \underline{-0.13} & -- & 0.12 & \underline{0.10} & 0.65 & 0.56 & -- & 0.44 \\
\texttt{Punct-diff} & -0.03 & 0.21 & -0.30 & -- & 0.18 & 0.07 & 0.43 & 0.44 & -- & 0.31 \\
\texttt{Entropy-diff} & -0.04 & 0.26 & -0.32 & -- & 0.21 & 0.01 & 0.47 & 0.45 & -- & 0.31 \\
\hline
\texttt{LaBSE} & \textbf{0.07} & \textbf{0.41} & \textbf{-0.57} & -- & \textbf{0.35} & \textbf{0.11} & \textbf{0.76} & \textbf{0.85} & -- & \textbf{0.57} \\
\hline
\texttt{Wegmann} (1) & -0.23 & -0.08 & \textbf{-0.00} & -- & 0.10 & \underline{0.18} & -0.43 & 0.10 & -- & -0.05 \\
\texttt{StyleDistance} (1) & -0.26 & -0.05 & \underline{-0.03} & -- & 0.12 & \textbf{0.29} & -0.43 & 0.00 & -- & -0.05 \\
\texttt{mStyleDistance} (3) & \underline{-0.00} & \underline{0.23} & -0.28 & -- & \underline{0.17} & -0.10 & \underline{0.50} & \underline{0.51} & -- & \underline{0.30} \\
\texttt{mStyleRepr} (3) & \textbf{0.01} & \textbf{0.50} & -0.58 & -- & \textbf{0.36} & -0.11 & \textbf{0.83} & \textbf{0.80} & -- & \textbf{0.50} \\
\hline
\texttt{Aya-8B} & 0.11 & 0.05 & 0.10 & 0.12 & 0.09 & 0.28 & 0.09 & 0.20 & 0.37 & 0.23 \\
\texttt{Gemma4-E4B} & 0.09 & -0.36 & 0.70 & 0.38 & 0.38 & 0.07 & 0.83 & 0.95 & 0.47 & 0.58 \\
\texttt{Gemma4-12B} & \textbf{0.39} & \underline{0.50} & 0.72 & \underline{0.48} & \underline{0.52} & \underline{0.49} & \underline{0.91} & 0.96 & \underline{0.60} & \underline{0.74} \\
\texttt{Gemma4-31B} & \underline{0.38} & 0.47 & \textbf{0.75} & \textbf{0.49} & \textbf{0.52} & \textbf{0.52} & 0.88 & \textbf{0.97} & \textbf{0.64} & \textbf{0.75} \\
\texttt{Llama31-8B} & 0.08 & 0.40 & 0.08 & 0.32 & 0.22 & 0.02 & 0.86 & 0.38 & 0.44 & 0.43 \\
\texttt{Llama32-3B} & 0.11 & 0.15 & 0.03 & 0.19 & 0.12 & 0.20 & 0.27 & 0.32 & 0.18 & 0.24 \\
\texttt{Mistral-7B} & 0.16 & 0.32 & 0.21 & 0.32 & 0.25 & 0.25 & 0.77 & 0.65 & 0.39 & 0.52 \\
\texttt{Olmo3-7B} & 0.09 & -0.14 & 0.09 & 0.12 & 0.11 & 0.35 & 0.29 & 0.30 & 0.22 & 0.29 \\
\texttt{Phi4-14B} & 0.15 & 0.42 & 0.49 & 0.38 & 0.36 & 0.16 & 0.86 & 0.80 & 0.49 & 0.58 \\
\texttt{Qwen25-7B} & 0.17 & 0.38 & 0.18 & 0.38 & 0.28 & 0.30 & 0.80 & 0.39 & 0.47 & 0.49 \\
\texttt{Qwen35-27B} & 0.25 & \textbf{0.50} & \underline{0.72} & 0.48 & 0.49 & 0.44 & 0.90 & \underline{0.97} & 0.54 & 0.71 \\
\texttt{Qwen35-4B} & 0.27 & 0.47 & 0.40 & 0.42 & 0.39 & 0.40 & 0.84 & 0.89 & 0.47 & 0.65 \\
\texttt{Qwen35-9B} & 0.24 & 0.48 & 0.68 & 0.44 & 0.46 & 0.37 & \textbf{0.93} & 0.90 & 0.51 & 0.68 \\
\texttt{Qwen3-8B} & 0.05 & 0.44 & 0.21 & 0.35 & 0.26 & 0.15 & 0.84 & 0.93 & 0.42 & 0.58 \\
\hline
\texttt{ClaudeHaiku-4.5} & \underline{0.23} & 0.44 & 0.67 & 0.39 & 0.43 & \underline{0.22} & 0.86 & \textbf{0.99} & 0.41 & 0.62 \\
\texttt{ClaudeOpus-4.8} & 0.10 & 0.48 & \underline{0.76} & \underline{0.49} & 0.46 & 0.13 & 0.84 & 0.96 & \textbf{0.59} & 0.63 \\
\texttt{ClaudeSonnet-5} & 0.15 & \textbf{0.50} & \textbf{0.76} & 0.48 & \underline{0.47} & 0.17 & \underline{0.89} & \underline{0.97} & 0.51 & \underline{0.64} \\
\texttt{GeminiFlashLite-3.1} & \textbf{0.40} & \underline{0.49} & 0.73 & 0.48 & \textbf{0.53} & \textbf{0.52} & \textbf{0.90} & 0.97 & 0.56 & \textbf{0.74} \\
\texttt{GeminiPro-3.1} & 0.10 & 0.48 & 0.75 & 0.49 & 0.45 & 0.10 & 0.88 & 0.97 & \underline{0.56} & 0.63 \\
\texttt{GPT-5.4-mini} & 0.06 & 0.42 & 0.48 & 0.33 & 0.32 & 0.11 & 0.72 & 0.88 & 0.36 & 0.52 \\
\texttt{GPT-5.5} & 0.07 & 0.45 & 0.73 & \textbf{0.52} & 0.45 & 0.01 & 0.88 & 0.95 & 0.56 & 0.60 \\
\bottomrule
\end{tabular}%
}
\end{table}

\begin{table}[t]\centering\small\setlength{\tabcolsep}{4pt}
\caption{Meta-evaluation of metrics calibrated on chance prediction ($\Phi = 0$) and human ceiling  ($\Phi = 0$); see Section~\ref{subsec:meta-metric} for implementation details. We present the results for individual metrics covering string-based metrics, semantic embeddings, style embeddings, open-weight models, and proprietary models. We also provide the performance of various ensemble metrics at the end (denoted by $\rho$).}
\label{tab:q-single}
\begin{tabular}{lccccc}\toprule
Metric ($k$) & Style & Factual & Leaks & Usability & Overall \\\midrule
\texttt{Levenshtein} (10) & 0.39 & \textbf{0.36} & \underline{0.63} & -- & \underline{0.46} \\
\texttt{Char3-cos} (10) & \underline{0.41} & \underline{0.34} & 0.59 & -- & 0.44 \\
\texttt{Verbatim} (10) & \textbf{0.43} & 0.32 & \textbf{0.69} & -- & \textbf{0.48} \\
\texttt{Length-ratio} (10) & 0.03 & 0.13 & 0.05 & -- & 0.07 \\
\texttt{Punct-diff} (10) & 0.06 & 0.11 & 0.15 & -- & 0.11 \\
\texttt{Entropy-diff} (10) & 0.13 & 0.16 & 0.18 & -- & 0.16 \\
\hline
\texttt{LaBSE} (10) & \textbf{0.43} & \textbf{0.35} & \textbf{0.73} & -- & \textbf{0.50} \\
\hline
\texttt{Wegmann} (1) & 0.02 & 0.00 & 0.00 & -- & 0.00 \\
\texttt{StyleDistance} (1) & 0.03 & 0.00 & 0.00 & -- & 0.01 \\
\texttt{mStyleDistance} (3) & \underline{0.07} & \underline{0.21} & \underline{0.17} & -- & \underline{0.15} \\
\texttt{mStyleRepr} (3) & \textbf{0.59} & \textbf{0.57} & \textbf{0.68} & -- & \textbf{0.61} \\
\hline
\texttt{Aya-8B} (10) & 0.07 & 0.06 & 0.02 & 0.07 & 0.05 \\
\texttt{Gemma4-E4B} (10) & 0.15 & 0.19 & 0.50 & 0.38 & 0.31 \\
\texttt{Gemma4-12B} (10) & \textbf{0.32} & \underline{0.41} & \underline{0.63} & \underline{0.43} & \underline{0.45} \\
\texttt{Gemma4-31B} (10) & \underline{0.31} & \textbf{0.43} & \textbf{0.66} & \textbf{0.47} & \textbf{0.47} \\
\texttt{Llama31-8B} (6) & 0.01 & 0.20 & 0.01 & 0.16 & 0.10 \\
\texttt{Llama32-3B} (5) & 0.02 & 0.01 & 0.02 & 0.03 & 0.02 \\
\texttt{Mistral-7B} (10) & 0.07 & 0.22 & 0.06 & 0.22 & 0.14 \\
\texttt{Olmo3-7B} (1) & 0.07 & 0.04 & 0.00 & 0.09 & 0.05 \\
\texttt{Phi4-14B} (10) & 0.07 & 0.39 & 0.20 & 0.32 & 0.24 \\
\texttt{Qwen25-7B} (10) & 0.12 & 0.32 & 0.05 & 0.31 & 0.20 \\
\texttt{Qwen35-27B} (10) & 0.16 & 0.40 & 0.52 & 0.34 & 0.36 \\
\texttt{Qwen35-4B} (10) & 0.11 & 0.38 & 0.17 & 0.28 & 0.24 \\
\texttt{Qwen35-9B} (10) & 0.06 & 0.34 & 0.28 & 0.31 & 0.25 \\
\texttt{Qwen3-8B} (10) & 0.08 & 0.38 & 0.06 & 0.42 & 0.23 \\
\hline
\texttt{ClaudeHaiku-4.5} (10) & 0.12 & \underline{0.43} & 0.54 & 0.39 & 0.37 \\
\texttt{ClaudeOpus-4.8} (10) & 0.24 & 0.42 & \textbf{0.68} & \textbf{0.55} & \underline{0.47} \\
\texttt{ClaudeSonnet-5} (10) & 0.18 & 0.43 & \underline{0.68} & \underline{0.51} & 0.45 \\
\texttt{GeminiFlashLite-3.1} (10) & \underline{0.30} & 0.40 & 0.61 & 0.45 & 0.44 \\
\texttt{GeminiPro-3.1} (10) & 0.22 & 0.41 & 0.63 & 0.38 & 0.41 \\
\texttt{GPT-5.4-mini} (10) & 0.10 & 0.33 & 0.30 & 0.25 & 0.25 \\
\texttt{GPT-5.5} (10) & \textbf{0.37} & \textbf{0.50} & 0.66 & 0.48 & \textbf{0.50} \\
\midrule\multicolumn{6}{l}{\emph{Meta-metrics on metric families (all covered languages)}}\\
$\rho_{string}$ & 0.50 & 0.42 & 0.70 & -- & 0.54 \\
$\rho_{embedding}$ & 0.48 & 0.36 & 0.74 & -- & 0.53 \\
$\rho_{string+embedding}$ & 0.58 & 0.44 & 0.74 & -- & 0.59 \\
$\rho_{open\text{-}source}$ & 0.54 & 0.55 & 0.69 & 0.60 & 0.59 \\
$\rho_{closed\text{-}source}$ & 0.69 & 0.52 & 0.69 & 0.68 & 0.65 \\
$\rho_{LLM}$ & 0.71 & 0.58 & 0.70 & 0.73 & 0.68 \\
$\rho_{all}$ & 0.78 & 0.61 & 0.79 & 0.73 & 0.73 \\
\bottomrule\end{tabular}\end{table}

\begin{table*}[t]\centering\scriptsize\setlength{\tabcolsep}{2.5pt}
\caption{Metric quality $\mathbf{\Psi}$ per language, for every single metric (top) and each ensemble of metrics (bottom). Each cell lists the scores for style, factual consistency, leakage, and usability (S/F/L/U). A score of 0 means the metric does no better than always predicting the mean label, and 1 means it matches the consensus as closely as a single annotator does; values are clipped to $[0,1]$. `--' marks dimensions a metric cannot score (usability has no reference text) and languages a metric does not cover.}
\label{tab:q-perlang}
\resizebox{\linewidth}{!}{%
\begin{tabular}{lcccccccccc}\toprule
Metric & \texttt{bn} & \texttt{el} & \texttt{en} & \texttt{fil} & \texttt{fr} & \texttt{fy} & \texttt{ja} & \texttt{ko} & \texttt{sl} & \texttt{sw} \\
& \scriptsize S/F/L/U & \scriptsize S/F/L/U & \scriptsize S/F/L/U & \scriptsize S/F/L/U & \scriptsize S/F/L/U & \scriptsize S/F/L/U & \scriptsize S/F/L/U & \scriptsize S/F/L/U & \scriptsize S/F/L/U & \scriptsize S/F/L/U \\\midrule
\texttt{Levenshtein} & .11/.12/.64/-- & .54/.84/.81/-- & .00/.07/.00/-- & .74/.38/.80/-- & .16/.02/.47/-- & .44/.25/.75/-- & .68/.77/.91/-- & .19/.06/.18/-- & .44/.63/.91/-- & .56/.47/.78/-- \\
\texttt{Char3-cos} & .07/.14/.64/-- & .57/.87/.83/-- & .00/.03/.00/-- & .80/.35/.69/-- & .16/.10/.58/-- & .83/.23/.56/-- & .69/.80/.87/-- & .16/.08/.17/-- & .30/.34/.79/-- & .54/.42/.75/-- \\
\texttt{Verbatim} & .11/.12/.67/-- & .65/.95/.89/-- & .00/.03/.00/-- & .85/.45/.90/-- & .16/.04/.74/-- & .84/.20/.77/-- & .63/.45/.97/-- & .17/.13/.30/-- & .33/.34/.88/-- & .58/.47/.81/-- \\
\texttt{Length-ratio} & .00/.03/.05/-- & .00/.29/.01/-- & .00/.00/.01/-- & .07/.01/.04/-- & .00/.04/.05/-- & .00/.25/.07/-- & .16/.53/.11/-- & .05/.02/.03/-- & .04/.06/.11/-- & .07/.04/.04/-- \\
\texttt{Punct-diff} & .00/.03/.19/-- & .13/.19/.19/-- & .00/.00/.00/-- & .00/.12/.07/-- & .00/.00/.16/-- & .00/.02/.15/-- & .41/.37/.14/-- & .01/.08/.09/-- & .04/.08/.34/-- & .04/.23/.19/-- \\
\texttt{Entropy-diff} & .00/.03/.06/-- & .35/.34/.47/-- & .01/.02/.04/-- & .12/.10/.28/-- & .02/.00/.14/-- & .14/.12/.17/-- & .34/.78/.25/-- & .00/.02/.13/-- & .31/.06/.12/-- & .01/.18/.15/-- \\
\hline
\texttt{LaBSE} & .15/.09/.77/-- & .68/.93/.93/-- & .00/.05/.00/-- & 1.0/.54/.96/-- & .17/.15/.78/-- & .46/.11/.88/-- & .54/.59/.95/-- & .28/.16/.34/-- & .37/.36/.88/-- & .54/.49/.78/-- \\
\hline
\texttt{Wegmann} & -- & -- & .02/.00/.00/-- & -- & -- & -- & -- & -- & -- & -- \\
\texttt{StyleDistance} & -- & -- & .03/.00/.00/-- & -- & -- & -- & -- & -- & -- & -- \\
\texttt{mStyleDistance} & -- & -- & -- & -- & .04/.00/.15/-- & -- & .10/.51/.34/-- & .08/.13/.03/-- & -- & -- \\
\texttt{mStyleRepr} & -- & .72/.82/.76/-- & -- & -- & .18/.18/.56/-- & -- & .86/.71/.73/-- & -- & -- & -- \\
\hline
\texttt{Aya-8B} & .00/.00/.00/.00 & .00/.05/.11/.00 & .00/.04/.00/.06 & .06/.00/.02/.00 & .00/.01/.03/.02 & .52/.41/.00/.37 & .05/.09/.03/.15 & .00/.00/.02/.00 & .11/.03/.00/.08 & .02/.02/.00/.08 \\
\texttt{Gemma4-E4B} & .20/.21/.49/.38 & .06/.20/.66/.53 & .00/.03/.22/.04 & .23/.20/.51/.23 & .01/.17/.60/.27 & .42/.16/.39/.67 & .28/.33/.68/.50 & .13/.12/.31/.18 & .15/.21/.61/.51 & .04/.27/.59/.51 \\
\texttt{Gemma4-12B} & .27/.10/.65/.36 & .18/.95/.75/.43 & .00/.01/.17/.00 & 1.0/.54/.61/.47 & .19/.11/.64/.36 & .00/.21/.67/.63 & .60/.46/.83/.57 & .34/.09/.31/.31 & .18/.92/.82/.63 & .11/.76/.82/.53 \\
\texttt{Gemma4-31B} & .29/.23/.72/.33 & .16/1.0/.79/.38 & .00/.02/.14/.01 & 1.0/.52/.59/.53 & .09/.11/.72/.43 & .03/.16/.64/.82 & .63/.50/.93/.83 & .32/.10/.36/.28 & .14/.92/.89/.53 & .20/.72/.82/.53 \\
\texttt{Llama31-8B} & -- & --/.42/.08/-- & .00/.06/.00/.07 & --/--/--/.10 & .01/.08/.00/.22 & .03/.21/.00/.49 & --/.23/.00/.05 & .01/.04/--/.06 & .04/.26/.01/.13 & .01/.33/.00/.19 \\
\texttt{Llama32-3B} & --/--/.03/-- & --/--/.04/-- & .00/.00/.00/.00 & .04/.01/.00/.06 & .00/.01/.03/.03 & -- & --/--/.01/-- & .02/.01/.02/.00 & .02/.00/.01/.10 & .07/.01/.02/-- \\
\texttt{Mistral-7B} & .00/.00/.00/.00 & .00/.06/.14/.00 & .01/.05/.00/.05 & .00/.21/.09/.22 & .00/.11/.23/.14 & .25/.67/.01/.55 & .08/.69/.01/.32 & .00/.00/.00/.04 & .30/.10/.07/.56 & .06/.31/.06/.34 \\
\texttt{Olmo3-7B} & --/.02/--/.00 & --/.00/--/.00 & .00/.02/.00/.00 & -- & -- & .31/.05/--/.38 & .10/.10/--/.13 & --/.02/--/.00 & .00/.00/--/.13 & .00/.10/--/.20 \\
\texttt{Phi4-14B} & .06/.02/.19/.17 & .00/.87/.36/.43 & .00/.04/.07/.06 & .24/.56/.26/.31 & .07/.06/.31/.35 & .11/.14/.21/.74 & .25/.51/.24/.53 & .00/.23/.06/.23 & .00/.85/.16/.16 & .05/.61/.15/.19 \\
\texttt{Qwen25-7B} & .10/.11/.08/.25 & .13/.59/.00/.53 & .01/.06/.03/.08 & .29/.38/.03/.22 & .10/.12/.19/.30 & .00/.12/.00/.35 & .01/.46/.13/.49 & .10/.09/.03/.14 & .31/.80/.04/.48 & .18/.45/.02/.28 \\
\texttt{Qwen35-27B} & .17/.05/.44/.29 & .16/1.0/.55/.30 & .01/.07/.24/.07 & .67/.53/.36/.27 & .08/.08/.63/.37 & .00/.16/.47/.46 & .35/.46/.74/.20 & .19/.11/.16/.30 & .00/.87/.87/.67 & .01/.68/.74/.51 \\
\texttt{Qwen35-4B} & .15/.04/.13/.05 & .05/.86/.18/.54 & .00/.13/.10/.00 & .56/.44/.11/.13 & .08/.16/.32/.31 & .00/.14/.09/.24 & .03/.47/.28/.39 & .06/.16/.19/.16 & .11/.83/.28/.63 & .10/.56/.07/.38 \\
\texttt{Qwen35-9B} & .07/.06/.23/.26 & .00/.83/.29/.43 & .00/.12/.11/.04 & .38/.46/.34/.31 & .07/.05/.41/.29 & .02/.10/.17/.26 & .00/.32/.20/.19 & .10/.11/.15/.20 & .00/.79/.50/.71 & .00/.52/.37/.42 \\
\texttt{Qwen3-8B} & .04/.04/.03/.34 & .00/.80/.07/.46 & .00/.08/.02/.03 & .10/.51/.04/.45 & .02/.11/.14/.43 & .37/.14/.04/.87 & .25/.36/.09/.40 & .00/.21/.10/.29 & .11/.91/.07/.42 & .00/.60/.01/.49 \\
\hline
\texttt{ClaudeHaiku-4.5} & .12/.06/.50/.28 & .02/1.0/.46/.47 & .01/.06/.30/.15 & .33/.55/.46/.24 & .08/.21/.64/.28 & .19/.18/.56/.63 & .18/.48/.79/.48 & .09/.21/.34/.19 & .14/.85/.70/.67 & .06/.72/.62/.53 \\
\texttt{ClaudeOpus-4.8} & .12/.10/.69/.46 & .23/.98/.86/.73 & .00/.02/.27/.13 & .11/.55/.62/.60 & .00/.12/.68/.41 & .25/.19/.64/.90 & 1.0/.53/.96/.76 & .11/.17/.39/.43 & .28/.90/.84/.53 & .26/.69/.82/.49 \\
\texttt{ClaudeSonnet-5} & .10/.04/.65/.44 & .31/1.0/.88/.63 & .00/.19/.28/.04 & .24/.55/.67/.51 & .15/.13/.70/.41 & .40/.21/.62/.92 & .13/.50/.92/.66 & .35/.09/.37/.51 & .09/.90/.84/.51 & .03/.70/.82/.49 \\
\texttt{GeminiFlashLite-3.1} & .34/.25/.69/.38 & .12/.77/.68/.53 & .03/.02/.13/.00 & 1.0/.53/.56/.48 & .17/.07/.67/.33 & .00/.16/.54/.86 & .48/.45/.86/.28 & .29/.14/.42/.40 & .17/.84/.79/.68 & .18/.74/.79/.55 \\
\texttt{GeminiPro-3.1} & .08/.20/.68/.44 & .06/.80/.67/.00 & .00/.04/.17/.00 & .11/.51/.59/.43 & .01/.06/.62/.36 & .19/.16/.61/.71 & 1.0/.54/.93/1.0 & .05/.16/.39/.24 & .22/.87/.79/.28 & .41/.72/.82/.31 \\
\texttt{GPT-5.4-mini} & .03/.09/.33/.19 & .03/.80/.40/.27 & .01/.05/.03/.05 & .00/.29/.27/.00 & .00/.09/.39/.28 & .59/.18/.28/.46 & .21/.41/.41/.34 & .05/.33/.13/.11 & .05/.52/.48/.48 & .05/.51/.30/.33 \\
\texttt{GPT-5.5} & .18/.24/.63/.37 & .40/1.0/.78/.44 & .00/.10/.25/.09 & .63/.56/.55/.56 & .05/.14/.69/.43 & .35/.17/.64/.58 & 1.0/.50/.95/.99 & .19/.66/.37/.34 & .35/.91/.89/.55 & .52/.73/.81/.43 \\
\midrule\multicolumn{11}{l}{\emph{Meta-metrics on metric families}}\\
$m_{string}$ & .11/.01/.65/-- & .59/.98/.88/-- & .00/.10/.05/-- & .83/.47/.89/-- & .15/.08/.73/-- & .93/.35/.77/-- & .84/.87/1.0/-- & .18/.08/.31/-- & .89/.68/.90/-- & .54/.56/.79/-- \\
$m_{embedding}$ & .15/.09/.77/-- & .81/.92/.93/-- & .00/.11/.01/-- & 1.0/.54/.96/-- & .22/.16/.79/-- & .46/.11/.88/-- & .77/.73/.95/-- & .26/.12/.41/-- & .37/.36/.88/-- & .54/.49/.78/-- \\
$m_{string+embedding}$ & .19/.02/.76/-- & .68/.99/.91/-- & .00/.10/.04/-- & 1.0/.54/.94/-- & .22/.27/.80/-- & .96/.35/.86/-- & .96/.88/1.0/-- & .30/.04/.42/-- & .88/.68/.89/-- & .54/.55/.79/-- \\
$m_{open\text{-}source}$ & .35/.21/.74/.48 & .39/1.0/.77/.80 & .00/.17/.23/.06 & 1.0/.55/.66/.71 & .21/.12/.73/.42 & .96/.74/.66/.96 & .88/.78/.95/.83 & .38/.18/.42/.37 & .60/.91/.89/.75 & .36/.79/.83/.62 \\
$m_{closed\text{-}source}$ & .41/.13/.69/.54 & .52/1.0/.86/.94 & .00/.19/.30/.17 & 1.0/.61/.67/.59 & .24/.19/.70/.53 & 1.0/.27/.62/.94 & 1.0/.63/.97/1.0 & .58/.49/.38/.56 & .63/.90/.89/.75 & .63/.74/.82/.62 \\
$m_{LLM}$ & .39/.20/.73/.54 & .59/1.0/.82/.92 & .00/.17/.27/.14 & 1.0/.62/.70/.88 & .28/.15/.72/.50 & 1.0/.74/.66/1.0 & 1.0/.79/.97/1.0 & .55/.40/.42/.57 & .73/.91/.89/.78 & .63/.79/.83/.64 \\
$m_{all}$ & .41/.30/.81/.54 & .81/1.0/.91/.92 & .00/.18/.32/.14 & 1.0/.58/.93/.88 & .30/.33/.84/.50 & 1.0/.68/.88/1.0 & 1.0/.99/1.0/1.0 & .64/.33/.45/.57 & .90/.90/.89/.78 & .67/.80/.81/.64 \\
\bottomrule
\end{tabular}%
}
\end{table*}

\begin{table*}[t]\centering\small\setlength{\tabcolsep}{3.5pt}
\caption{Percentage of LLM-judge predictions without a usable score, per language, pooled over the four evaluation dimensions. A prediction counts as failed if the output could not be parsed, was cut off before the score, or was empty or a refusal. Each cell lists pointwise\,/\,pairwise. Pairwise rates are based on far fewer items (20--40 per language and dimension, against about 200 pointwise).}
\label{tab:parse-errors-perlang}
\resizebox{\linewidth}{!}{%
\begin{tabular}{lcccccccccc|c}\toprule
Model & \texttt{bn} & \texttt{el} & \texttt{en} & \texttt{fil} & \texttt{fr} & \texttt{fy} & \texttt{ja} & \texttt{ko} & \texttt{sl} & \texttt{sw} & Overall \\
\texttt{Aya-8B} & 6/53 & 1/8 & 1/0 & 3/13 & 2/5 & 1/20 & 0/19 & 0/2 & 1/2 & 0/15 & 1/14 \\
\texttt{Gemma4-E4B} & 1/1 & 1/3 & 0/0 & 1/2 & 1/3 & 1/1 & 1/0 & 1/2 & 1/2 & 1/2 & 1/2 \\
\texttt{Gemma4-12B} & 0/1 & 0/1 & 0/0 & 0/2 & 0/3 & 0/1 & 0/1 & 0/2 & 0/1 & 0/6 & 0/2 \\
\texttt{Gemma4-31B} & 1/0 & 0/3 & 0/0 & 1/2 & 0/2 & 0/2 & 0/2 & 0/5 & 0/2 & 0/4 & 0/2 \\
\texttt{Llama31-8B} & 80/87 & 47/16 & 4/0 & 52/19 & 11/5 & 24/8 & 36/8 & 26/7 & 28/10 & 23/38 & 33/20 \\
\texttt{Llama32-3B} & 63/31 & 55/56 & 11/0 & 29/62 & 30/37 & 63/30 & 52/16 & 20/3 & 32/22 & 42/53 & 40/35 \\
\texttt{Mistral-7B} & 2/25 & 0/21 & 0/0 & 0/0 & 0/0 & 1/0 & 0/0 & 0/0 & 0/0 & 0/0 & 1/5 \\
\texttt{Olmo3-7B} & 40/65 & 45/56 & 12/37 & 67/27 & 60/56 & 21/15 & 20/38 & 56/23 & 23/18 & 21/8 & 36/34 \\
\texttt{Phi4-14B} & 12/53 & 7/28 & 3/0 & 7/4 & 4/8 & 9/3 & 7/13 & 8/5 & 7/4 & 7/6 & 7/13 \\
\texttt{Qwen2.5-7B} & 2/9 & 2/3 & 0/0 & 2/0 & 1/0 & 1/0 & 0/0 & 0/0 & 0/0 & 1/0 & 1/1 \\
\texttt{Qwen3.5-27B} & 5/4 & 3/14 & 2/3 & 2/7 & 2/5 & 6/6 & 1/17 & 3/17 & 1/10 & 2/5 & 3/9 \\
\texttt{Qwen3.5-4B} & 10/25 & 28/8 & 2/0 & 14/4 & 3/7 & 12/3 & 2/1 & 2/7 & 4/4 & 15/5 & 9/7 \\
\texttt{Qwen3.5-9B} & 12/37 & 14/22 & 7/32 & 10/17 & 9/20 & 15/11 & 12/26 & 11/43 & 9/8 & 14/23 & 11/22 \\
\texttt{Qwen3-8B} & 10/15 & 11/10 & 0/0 & 1/1 & 1/0 & 2/0 & 5/1 & 2/1 & 2/0 & 3/0 & 4/3 \\
\hline
\texttt{ClaudeHaiku-4.5} & 0/0 & 0/0 & 1/5 & 0/0 & 0/0 & 0/0 & 0/1 & 0/0 & 0/0 & 0/0 & 0/0 \\
\texttt{ClaudeOpus-4.8} & 0/0 & 0/0 & 0/0 & 1/0 & 0/0 & 0/0 & 0/0 & 0/0 & 12/5 & 14/36 & 3/4 \\
\texttt{ClaudeSonnet-5} & 1/3 & 2/5 & 5/29 & 1/0 & 0/3 & 3/0 & 1/3 & 1/8 & 12/11 & 14/39 & 4/8 \\
\texttt{GeminiFlashLite-3.1} & 2/0 & 16/9 & 0/0 & 2/1 & 4/3 & 1/2 & 0/0 & 1/1 & 4/7 & 5/3 & 3/3 \\
\texttt{GeminiPro-3.1} & 2/0 & 18/12 & 0/3 & 4/2 & 4/4 & 7/2 & 3/0 & 3/2 & 6/7 & 9/3 & 6/4 \\
\texttt{GPT-5.4-mini} & 0/0 & 0/0 & 0/0 & 0/0 & 0/0 & 0/1 & 0/0 & 0/0 & 0/0 & 0/0 & 0/0 \\
\texttt{GPT-5.5} & 0/0 & 0/0 & 0/0 & 0/0 & 0/0 & 0/0 & 0/0 & 0/0 & 0/0 & 0/0 & 0/0 \\
\bottomrule\end{tabular}%
}
\end{table*}

\section{Prompts} \label{appx:prompt}

\begin{figure}
    \centering
    \includegraphics[width=0.85\linewidth]{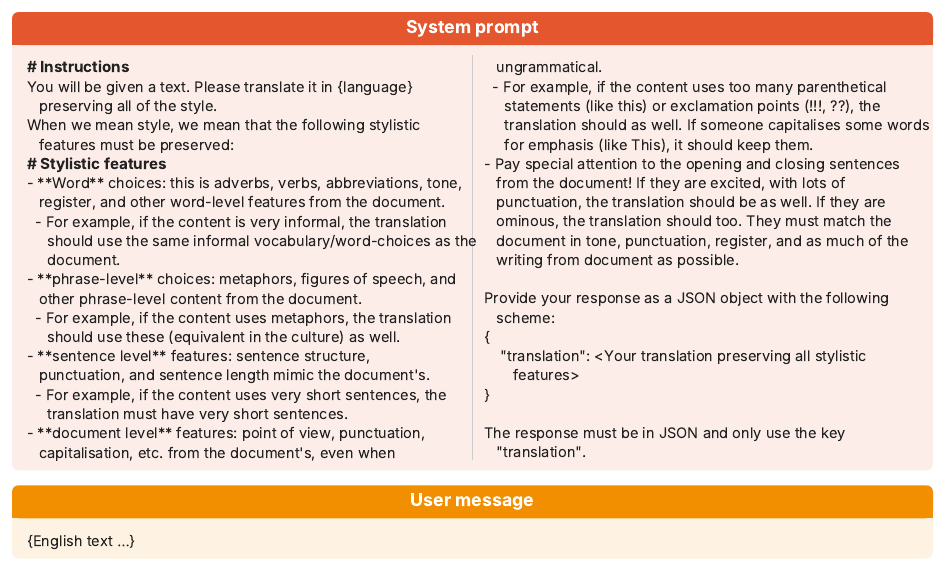}
    \caption{The zero-shot prompt used to translate English text to other languages.}
    \label{fig:prompt-translate}
\end{figure}

\begin{figure}
    \centering
    \includegraphics[width=0.85\linewidth]{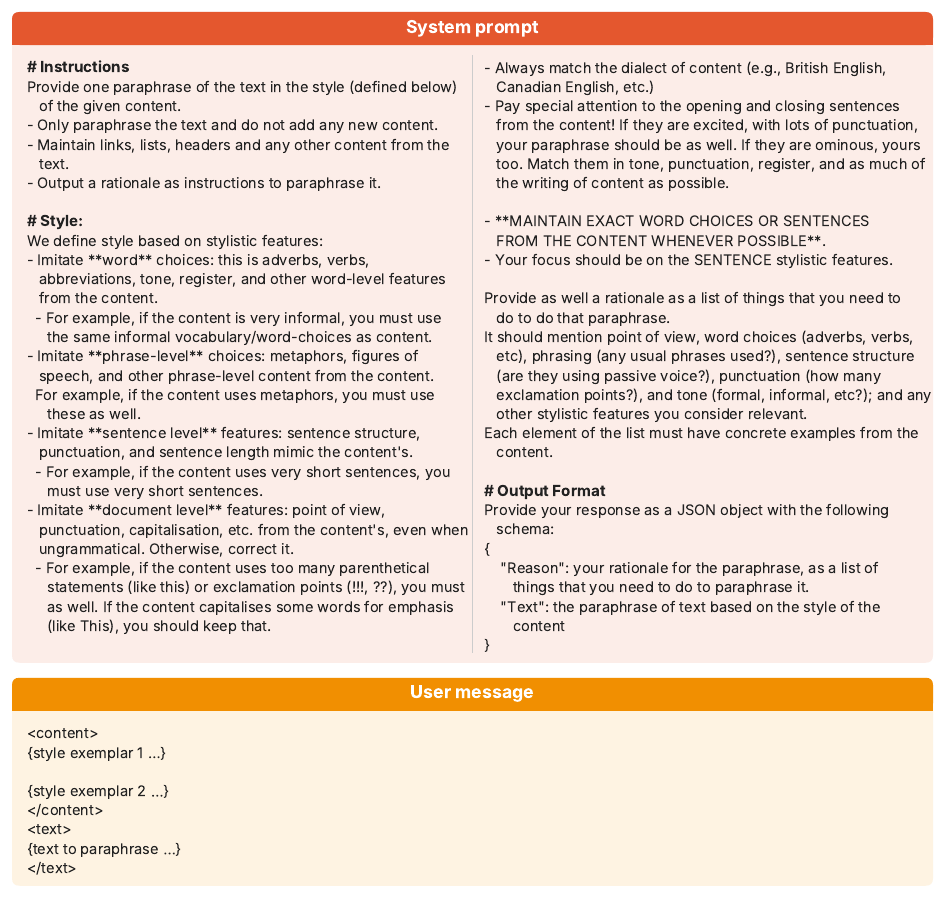}
    \caption{The zero-shot prompt used to generate style-personalized text for all languages.}
    \label{fig:prompt-sptg}
\end{figure}

\begin{figure}
    \centering
    \includegraphics[width=0.85\linewidth]{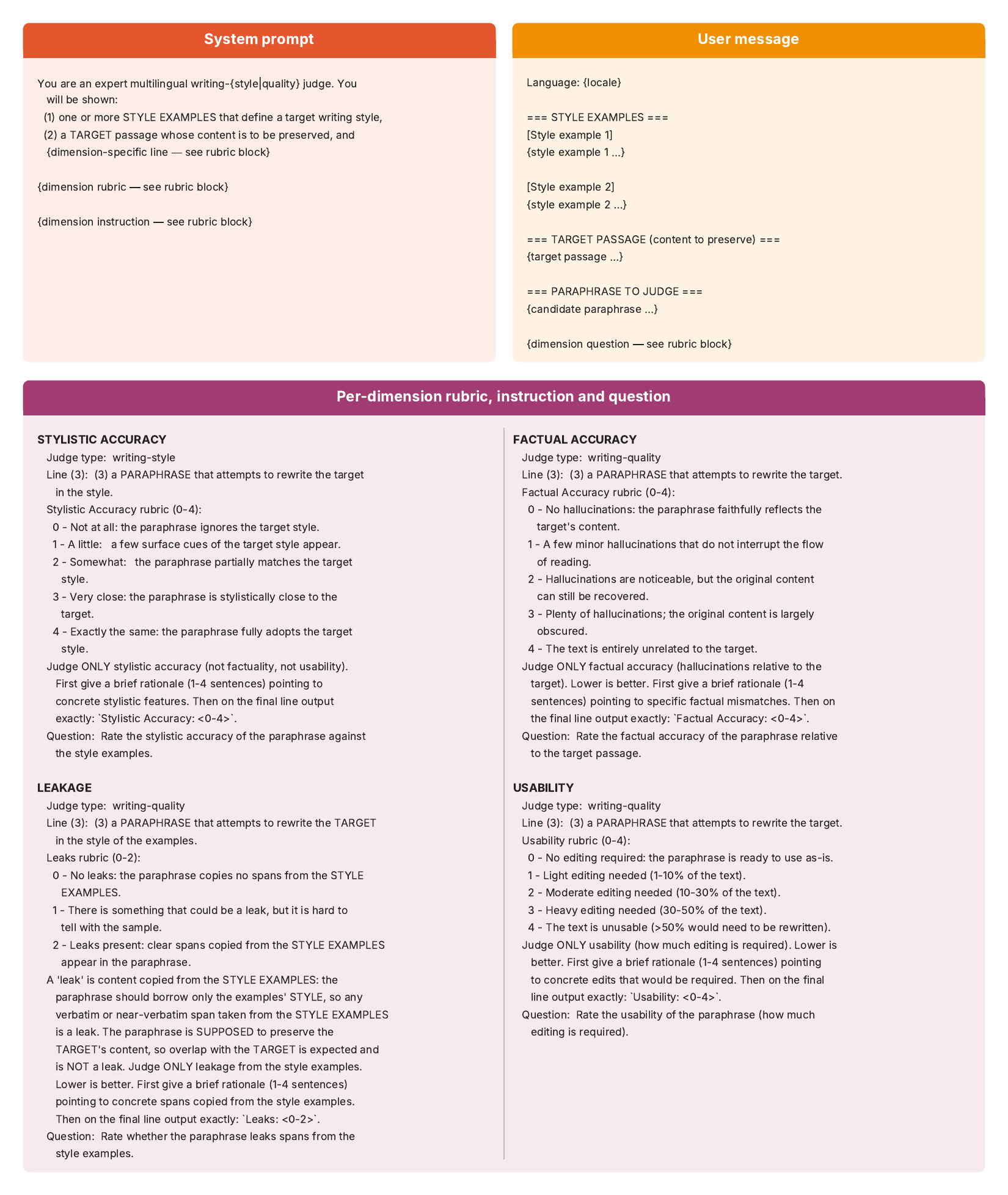}
    \caption{The zero-shot LLM-as-a-judge prompt used to obtain the pointwise LLM responses.}
    \label{fig:judge-prompt}
\end{figure}

\begin{figure}
    \centering
    \includegraphics[width=0.85\linewidth]{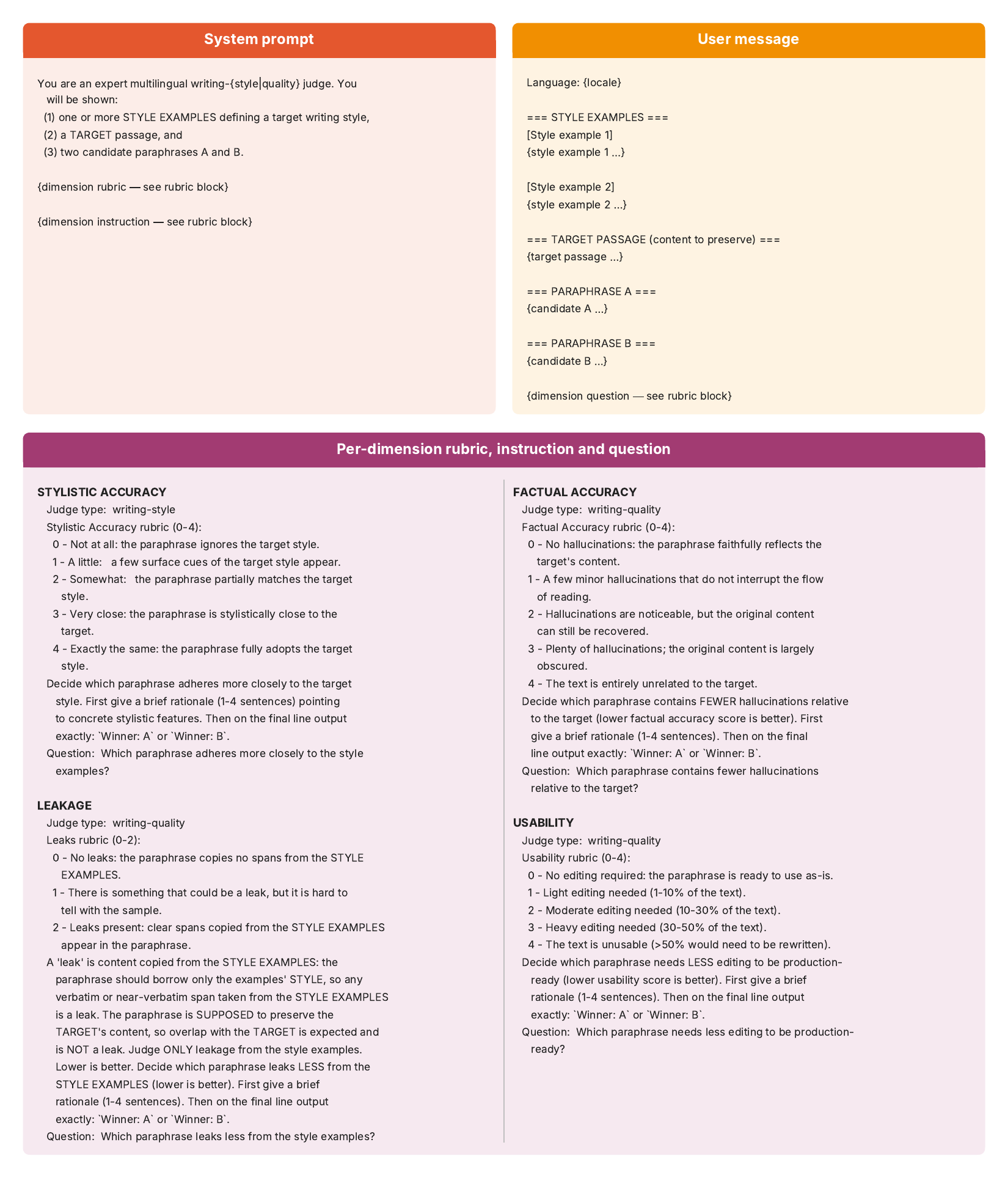}
    \caption{The zero-shot LLM-as-a-judge prompt used to obtain the pairwise LLM responses.}
    \label{fig:judge-prompt-pairwise}
\end{figure}

Figures~\ref{fig:prompt-translate} and~\ref{fig:prompt-sptg} provide the dataset construction prompts for translating English text into other languages and generating SPTG output, respectively.
Figures~\ref{fig:judge-prompt} and~\ref{fig:judge-prompt-pairwise} provide the pointwise and pairwise evaluation prompts used to generate the LLM-as-judge responses for all the judge models.

\end{document}